\documentclass{article}  

\usepackage{geometry}
\usepackage[utf8]{inputenc} % allow utf-8 input
\usepackage[T1]{fontenc}    % use 8-bit T1 fonts
\usepackage{hyperref}       % hyperlinks
\usepackage{url}            % simple URL typesetting
\usepackage{booktabs}       % professional-quality tables
\usepackage{amsfonts}       % blackboard math symbols
\usepackage{nicefrac}       % compact symbols for 1/2, etc.
\usepackage{microtype}      % microtypography
\usepackage{xcolor}         % colors

\usepackage{authblk}
\usepackage{graphicx}
\usepackage{booktabs}
\usepackage{longtable}
\usepackage{tabu}
\usepackage{listings}
\usepackage{amsthm}
\usepackage{amsmath}
\usepackage{colortbl}
\usepackage{subcaption}

\usepackage{pythonhighlight}
\usepackage{dirtytalk}

\title{Continual learning in Deep Neural Networks:\\ an Analysis of the Last Layer}

\author{Timoth\'ee Lesort}
\author{Thomas George}
\author{Irina Rish}
\affil{Université de Montréal, MILA - Quebec AI Institute}

\begin{document}

\maketitle

\begin{abstract}
We study how different output layer parameterizations of a  deep neural network affects learning and forgetting in continual learning settings. The following three effects can cause catastrophic forgetting in the output layer: (1) weights modifications, (2) interference, and (3) projection drift.
In this paper, our goal is to provide more insights into how changing the output layer parameterization may address (1) and (2). 
Some potential solutions to those issues are proposed and evaluated here in several continual learning scenarios.
We show that the best-performing type of output layer depends on the data distribution drifts and/or the amount of data available. In particular, in some cases where a standard linear layer would fail, changing parameterization is sufficient to achieve a significantly better performance, without introducing any continual-learning algorithm but instead by using standard SGD to train a model. 
Our analysis and results shed light on the dynamics of the output layer in continual learning scenarios and suggest a way of selecting the best type of output layer for a given scenario.

\end{abstract}

\section{Introduction}
Continual deep learning algorithms usually rely on end-to-end training of deep neural networks, making them difficult to analyze given the complexity (non-linearity, non-convexity) of the training dynamics. 
In this work, we instead propose to evaluate the ability of a network to learn or forget in a simplified learning setting: 
We decompose our model as a \emph{feature extractor} part consisting of all but the final layer, and a \emph{classifier} part which is given by the output layer. Using this decomposition, we can isolate the role of the output layer parameterization in continual learning scenarios.

Recent works on continual learning (CL) point out some drawbacks of parameterizing the output layer as a linear layer, especially in incremental settings \cite{wu2019large,zhao2020maintaining,Hou_2019_CVPR}. Among these problems is the unbalance of bias and/or norm of the output layer's vectors. These works propose some solutions but do not study them independently from the feature extractor. Indeed, in CL scenarios, the function learned by the feature extractor changes, and therefore the embedding space changes accordingly; we call this change the \textit{projection drift}. This drift characterizes the change through time of the embedding on a given observation.  

By isolating the output layer from the feature extractor, we can study it in a controlled environment, i.e., without projection drifts.
Decoupling the feature extractor from a sub-network has shown to be helpful, e.g., in reinforcement learning  \cite{Lesort18,raffin2019decoupling,stooke2020decoupling}. 

We now list our contributions:\\
%
%\begin{itemize}
$\bullet$ We propose an evaluation of a large panel of output layer types in incremental, lifelong, and mixed continual scenarios. We show that depending on the type of scenario, the best output layer may vary.

$\bullet$  We describe the different sources of performance decrease in continual learning for the output layer: forgetting, interference, and projection drifts.

$\bullet$  We propose different solutions to address catastrophic forgetting (CF) in the output layer: a simplified weight normalization layer, and two masking strategies Sec. \ref{sec:layers}.

$\bullet$ We provide a proof of concept for end-to-end training by applying a reparameterized layer to a vanilla replay method Sec. \ref{sec:e2e}. 
%\end{itemize}

\section{Related Works}

In most deep continual learning papers, the last layer of the deep neural network is parameterized as a linear layer as in typical classifier architectures in i.i.d. settings such as VGG \cite{simonyan2014very}, ResNet \cite{he2016deep}, and GoogleNet \cite{szegedy2014going}.
Recently, several CL approaches questioned this approach, showing that the norm or the bias can be unbalanced for the last classes observed in class-incremental settings. For example, BIC \cite{wu2019large} proposed to train two additional parameters using a subset of the training dataset after training the rest of the model to correct for the unbalance of the bias. In the same spirit, \cite{zhao2020maintaining} proposed to compute a \textit{Weight Aligning} value to apply to vector to balance the norm of past classes vectors and the new classes vectors of a linear layer. \cite{zhao2020maintaining} also experiments with a layer, the weight normalization layer \cite{salimans2016weight}, that decouples the norm of the weight matrix from its direction to normalize the vector of linear output layers.
\cite{Hou_2019_CVPR,caccia2021reducing} proposed to apply a cosine normalization of the layer to avoid the norm and bias unbalance.  

% SLDA related papers
Linear Discriminant Analysis (LDA) Classifiers in incremental settings have been used for dozens of years \cite{ALIYARIGHASSABEH20151999,Chatterjee1997self,DEMIR2005421}.
The streaming version SLDA from \cite{PangIncremental2005} has recently been revisited in deep continual learning in \cite{hayes2020lifelong}. SLDA combines the mean and the covariance of the training data features to classify new observations. 
%
% NMC
A more straightforward approach, as implemented in iCaRL \cite{rebuffi2017icarl} is to average the features for each class and apply a nearest neighbor classifier also called Nearest Mean Classifier (NMC). 

In this work, we study the capability of those various types of top layers with a fixed pre-trained model in incremental and lifelong settings as defined in \cite{lesort2021understanding}.
Many different strategies have been proposed in the continual learning community: rehearsal, generative replay, dynamic architectures or regularization. In this study, we do not aim at modifying the training process or proposing a new one. We instead study how different layer types learn and forget in various continual scenarios without any additional CL approach.

\section{Output Layer Types}
\label{sec:layers}

We here review the output layer parameterizations studied in this work. We introduce the notations and present the main characteristics of linear layers. Then, we explain how the standard layer parameterization struggles to learn continually, and we present different parameterizations to overcome the problems. Finally, we present several other types of output layers not trained by gradient descent.

\subsection{Notations}

We will study functions $f_{\theta}(\cdot)$ parameterized by a vector of parameters $\theta$ representing the set of weight matrices and bias vectors of a deep network. In continual learning, the goal is to train $f_{\theta}(\cdot)$ on a sequence of task $[\mathcal{T}_0, \mathcal{T}_1, ..., \mathcal{T}_{T-1}]$, such that $\forall (x,y)\sim \mathcal{T}_t$ ($t \in [0, T-1]$), $f_{\theta}(x)=y$.

We can decompose $f_{\theta}(\cdot)$ into two parts, (1) a feature extractor $\phi_{\theta^-}(\cdot)$ which encodes the observation $x$ into a feature vector $z$, (2) an output layer $g_{\theta^+}(\cdot)$ which transforms the feature vector into a non-normalized vector $o$ (the logits). $\theta^-$ and $\theta^+$ are two complementary subsets of $\theta$. 

\subsection{Linear Layer}

 A linear layer is parameterized by a weight matrix $A$ and bias vector $b$, respectively of size $N \times h$ and $N$, where $h$ is the size of the latent vector (the activations of the penultimate layer) and $N$ is the number of classes.
For $z$ a latent vector, the output layer computes the operation $o = A z + b$.
We can formulate this operation for a single class $i$ with $\langle z, A_i \rangle + b_i = o_i$, where $\langle \cdot \rangle$ is the euclidean scalar product, $A_i$ is the $i$th row of the weight matrix viewed as a vector and $b_i$ is the corresponding scalar bias.

It can be rewritten:
\begin{equation}
\lVert z \rVert \lVert A_i \rVert \cdot cos(\angle(z, A_i)) + b_i = o_i
\label{eq:linear}
\end{equation}
Where $\angle(\cdot, \cdot)$ is the angle between two vectors and $\lVert \cdot \rVert$ denotes here the euclidean norm of a vector.

\begin{figure}[h]
    \centering
    \begin{subfigure}{0.32\columnwidth}
        \centering
        \includegraphics[width=\linewidth]{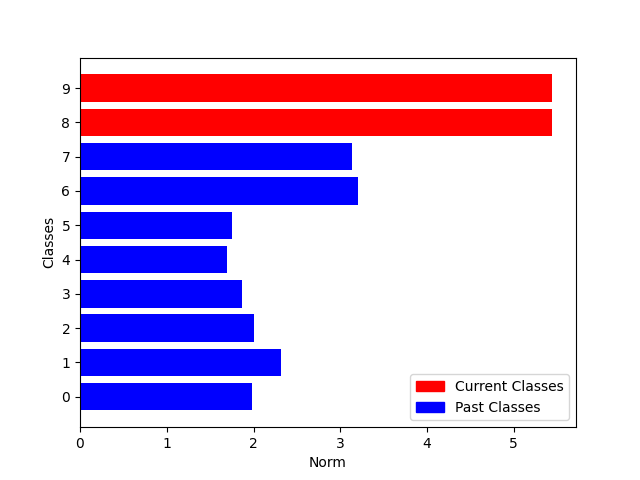}
    \end{subfigure}
    \begin{subfigure}{0.32\columnwidth}
        \centering
        \includegraphics[width=\linewidth]{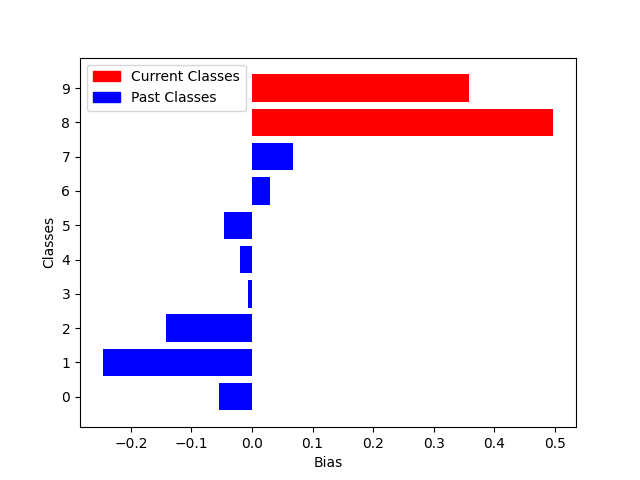}
    \end{subfigure}
    \begin{subfigure}{0.32\columnwidth}
        \centering
        \includegraphics[width=\linewidth]{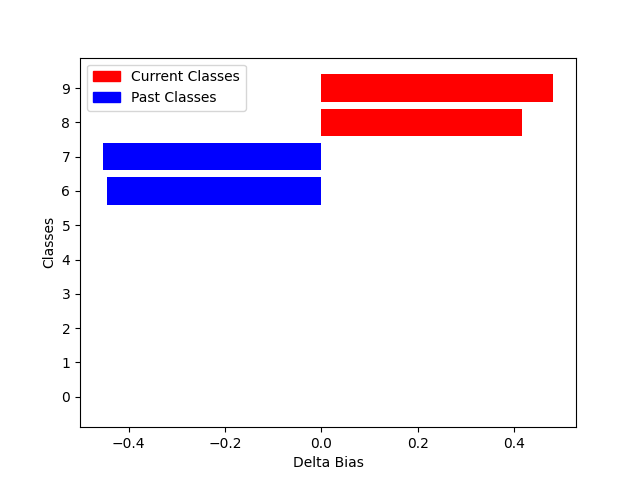}
    \end{subfigure}
    \caption{Illustration of norm and bias unbalance at the end of a CIFAR10 continual experiment. 5 tasks with 2 classes each. (left) the norm of each output vector at the end of the last task, (middle) bias at the end of the last task, and  (right) the difference between the last task's bias and current bias. We can see a clear unbalance in norm and bias for the last active vectors (classes 8 and 9). The middle right figure shows us that the imbalance of bias is primarily due to the modification of bias from the previous task (classes 6 and 7).
    }
    \label{fig:norm_unbalance_linear_cifar}
\end{figure}

Eq. \ref{eq:linear} highlights the 3 components that need to be learned to make a correct prediction $\hat{y}=\operatorname*{argmax}_i (o_i)$: The norm of $A_i$, its angle with the latent representation, and the bias. In particular, if the training dynamics make a particular $\lVert A_i \rVert$ or $b_i$ larger than others, then the network will be considerably biased into predicting the class $i$. 
%While this problem is not present in an i.i.d setting where all parameters are learned using all classes,
In incremental learning, the vectors $A_i$ and values $b_i$ are learned sequentially, hence the learning algorithm will grow the norm $\lVert A_i \rVert$ and bias $b_i$ for the current classes and reduce them for past classes to ease the current task (Figure \ref{fig:norm_unbalance_linear_cifar}).
This may lead to forgetting.

\subsection{Linear classifiers in continual learning and reparameterizations}
\label{sub:linear_cl}
%
% setting description
In continual learning, assuming that the optimal function $f(x)=y$ is fixed, there are two main cases of data distribution drift: % (see e.g. \cite{lesort2021understanding}):
either we get new examples from new classes (virtual concept drift $\xrightarrow[]{}$ incremental scenario) or from known classes but in slightly different setups (domain drift $\xrightarrow[]{}$ lifelong scenario). For a linear output layer, forgetting is caused by different mechanisms in both scenarios.

\begin{figure}[h]
    \centering
    \begin{subfigure}[t]{0.3\linewidth}
        \centering
        \includegraphics[width=\linewidth]{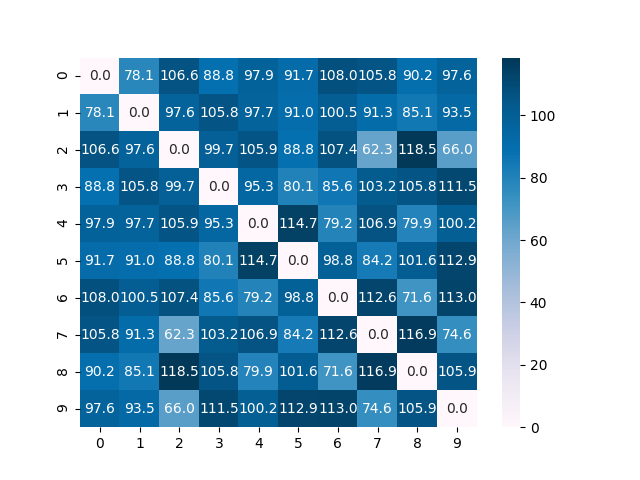}
        \label{fig:linear_vectors_angles}
        \caption{Angles between vectors}
    \end{subfigure}
    \begin{subfigure}[t]{0.3\linewidth}
        \centering
        \includegraphics[width=\linewidth]{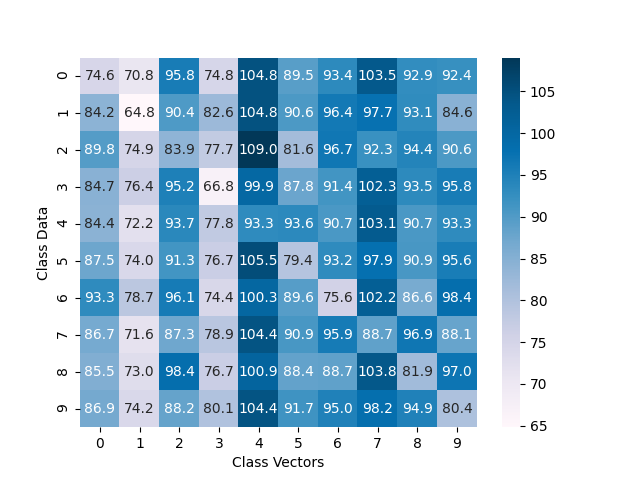}
        \label{fig:linear_vectors_angles_data}
        \caption{ Mean angles vectors-data.}
    \end{subfigure}
    \begin{subfigure}[t]{0.3\linewidth}
        \centering
        \includegraphics[width=\linewidth]{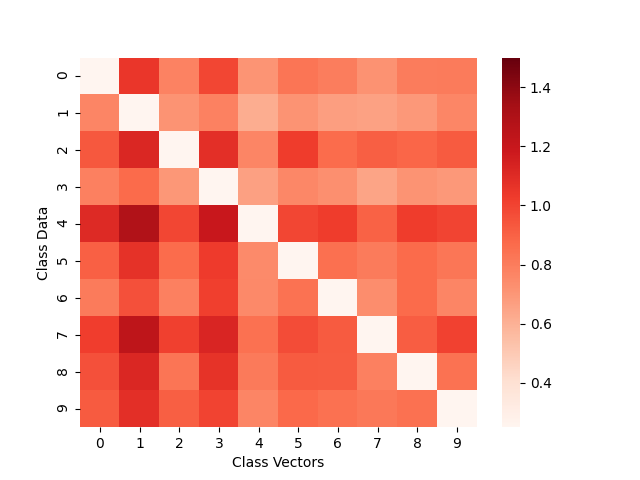}
        \label{fig:linear_interference_risks}
        \caption{Interference risk}
    \end{subfigure}
    \caption{Illustration of \emph{interference} for the linear layer: We plot \textbf{(left)} the angles (in degree) between the output vectors $A_i$, \textbf{(middle)} the mean angle between the latent vectors $z$ of each class and the $A_i$ vectors and \textbf{(right)} ratio between the mean angle with wrong classes, and the mean angle of the target class: a high value indicates a higher risk of interference. In this experiment, there is a high risk of interference between examples for class 4 and examples of class 1 or 3, as pictured by a dark red cell.
     %See appendix \ref{ap:visualization} for a similar experiment for WeightNorm.
     }
    \label{fig:interference}
\end{figure}

% problems
%\paragraph{Incremental: interferences}
\textbf{In incremental learning}, the $A_i$ and $b_i$ are learned sequentially (one by one or set by set). 
% -> tester les vecteurs appris dans le passé avec des nouveaux vecteurs
When learning a new $A_i$ (and $b_i$), the training algorithm does not have access to past classes, and hence cannot ensure that learning weight vectors for new classes does not interfere with past classes (eq. \ref{eq:linear}).
As an illustration, suppose two classes are alike in the latent space. In that case, their respective latent vectors might be similar, and \textbf{interference} might happen between the two classes  (see fig. \ref{fig:interference}), a fortiori if the norm and bias are not balanced. %Interference might be due to forgetting but not exclusively. 

\begin{figure}[h]

        \centering
    \begin{subfigure}{0.49\linewidth}
        \centering
            \includegraphics[width=\linewidth]{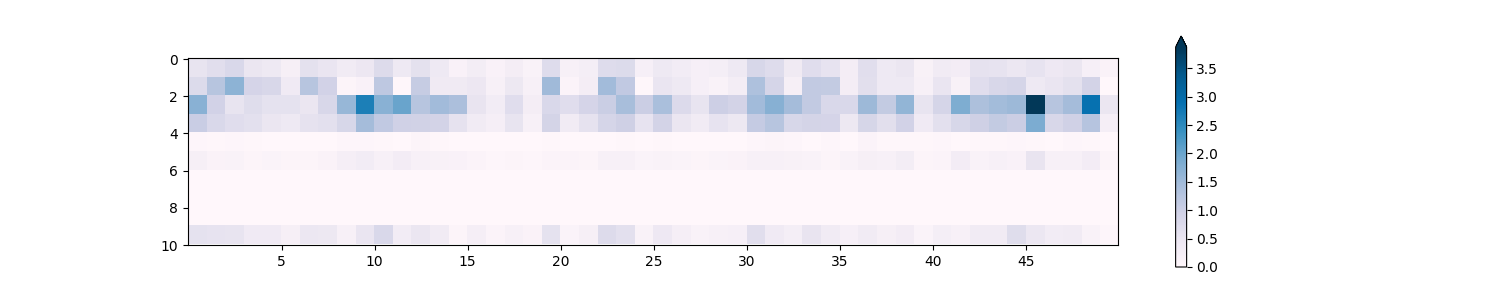}
        \caption{No masking}
    \end{subfigure}
    \begin{subfigure}{0.49\linewidth}
        \centering
            \includegraphics[width=\linewidth]{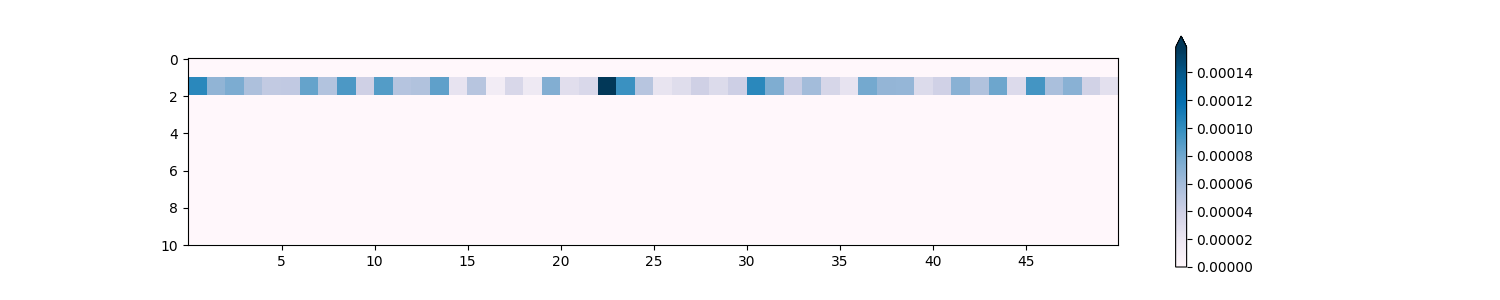}
        \caption{Single masking}
    \end{subfigure}
    
    \caption{ Illustration of Forgetting: we plot the difference between weights before and after task 45 in the scenario Core10Mix (first 50 dimensions only). Task 45 is composed of only one class that is different from 44's one. Those two figures show the impact of masking. Indeed, it avoids the modification of weights of other classes.
    }
    \label{fig:weight_modif}
\end{figure}

One solution found in the bibliography to avoid interference is to counteract unbalance between the norms of vectors or bias. We can modify eq. \ref{eq:linear} in several ways to mitigate such unbalance:\\
%
% potential solutions
Removing the bias (\textit{Linear\_no\_bias} layer):
\begin{equation}
   \lVert z \rVert \lVert A_i \rVert \cdot cos(\angle(z, A_i)) = o_i
\label{eq:linear_wo_bias}
\end{equation}
Normalizing output vectors (\textit{WeightNorm} layer):
\begin{equation}
  \lVert z \rVert \cdot cos(\angle(z, A_i)) = o_i
\label{eq:weightnorm}
\end{equation}
Measuring only the angle (\textit{CosLayer}):
\begin{equation}
 cos(\angle(z, A_i)) = o_i
\label{eq:coslayer}
\end{equation}
WeightNorm (eq. \ref{eq:weightnorm}) is similar to the original WeightNorm layer (\cite{salimans2016weight}, here denoted by \textit{Original WeightNorm}) experimented in \cite{zhao2020maintaining} in a continual learning context:
\begin{equation}
 \gamma_i \lVert z \rVert \cdot cos(\angle(z, A_i)) + b_i = o_i
\label{eq:original_weightnorm}
\end{equation}
However, in the original WeightNorm, the additional scaling parameter $\gamma$ and the bias $b$ are learned during training. These parameters are akin to the parameters in BatchNorm layers \cite{ioffe2015batch}, which have been shown to be more prone to catastrophic forgetting in the intermediate layers in continual learning \cite{lomonaco2020rehearsal}.
Hence, our proposed WeightNorm layer (eq. \ref{eq:weightnorm}) avoids such interference by ensuring a unit norm for all vectors and removing bias and gamma parameters.

\textbf{Forgetting might also be caused by weight modification}: this happens when the optimizer modifies weight vectors from past classes that are not found in the current task. In the literature, regularization strategies (e.g. EWC \cite{kirkpatrick2017overcoming})
%, KFRA \cite{Ritter18Online})
aims at avoiding such forgetting by penalizing modification of important parameters. 

We introduce a more radical strategy to avoid that updating a specific class affects another past one. The strategy consists in masking some classes during the update step. We propose two types of masking: in \textit{single masking}, we only update weights for the output vector of the true target and in \textit{group masking}, we mask all classes that are not in the mini-batch. With the following strategies, the update step cannot change $A_i$ and $b_i$ of past classes and avoid sub-sequential interference. 
In Figure \ref{fig:weight_modif}, we illustrate the impact of \textit{single masking} in a task with only one class from one experiment of the paper. In incremental scenarios, all masking strategies can also be seen as a regularization strategy that strictly forbids the modification of past weights in the last layer.

% comparison between masking and multi-head
The masking strategy resembles the update step in a multi-head architecture \cite{van2019three} for gradient descent since in this case, the gradient is computed only for a subset of selected classes. However, in our setup, there is only one head, and the inference is achieved without the help of any external supervision, such as a task label.  

\textbf{In lifelong learning}, forgetting happens when new instances of a previously learned class make the output layer forget features that were important for past instances or modify the direction of the output vectors. %This is not inter-classes interference but mostly intra-class interference. 
However, those problems are usually less prone to catastrophic forgetting (CF) but may still have a significant impact in complex scenarios. One of the reasons that CF is less important is that all classes are available simultaneously, which makes it possible for the output vectors to be coordinated and avoid interference.

\subsection{Classifiers Trained without Gradient}

To avoid interference and weight modification, we experiment with another family of layer types that rely on the similarity between training latent vectors and testing latent vectors but are not trained by gradient descent.

% KNN
One of the most popular classifiers that can easily be adapted into a continual or online learning setting is the k-nearest neighbors (KNN) classifier. This classifier searches for the k nearest exemplars of a given observation in the training set and predicts the class based on the class of the k neighbors. This classifier's hyper-parameters are the number of neighbors $k$ and the distance used (here, the usual euclidean norm). 

% NMC: nearest mean classifier
A lighter strategy is the nearest mean classifier (denoted  \textit{MeanLayer}) as proposed in iCarL \cite{rebuffi2017icarl}, this strategy consists of saving the mean of the features of each class $k$ while learning (called a \emph{prototype} $\mu_k$) and predict the class given by the nearest mean for test latent vectors.

% SLDA
An alternative is the linear discriminant analysis classifier, which uses both the mean of the features and a covariance. The online version of this algorithm is streaming linear discriminant analysis (SLDA) \cite{PangIncremental2005}.

While similarity-based classifiers have proven very effective in practice, they cannot be directly used to train the feature extractor since they are not differentiable. Hence, they need to store samples from past tasks as rehearsal strategies or be applied in a setting without projection drift.

\medskip

In this section, we introduced the different output layer types that we will use in experiments. We introduce two modifications: (1) a simplified WeightNorm (referred to as \textit{WeightNorm}), and, (2) two masking strategies, single masking and group masking, compatible with any reparameterization of the output layer.

\section{Experiments}
\label{sec:exps}

In this section, we will review how the different output layer types, presented in section \ref{sec:layers}, can solve various continual scenarios. We isolate the output layer from the remaining of the neural network by using a frozen pre-trained model. This methodology consists of creating a scenario where no projection drift occurs.  
Indeed, projection drifts may only lead to a deterioration of performance. Hence our study can eliminate approaches that would, in all cases, fail under such conditions. Our findings can also be directly transferred to settings where the projection drift is negligible. 
%

% introduction of all experiments
In a preliminary experiment, we benchmark all output layer types in an i.i.d. setting to fix some of their hyper-parameters for further continual training. The conclusion of these experiments are in Sec.~\ref{sub:data_model}. 
In a second setup, we train the various output layers in continual learning scenarios. Finally, we experiment with a vanilla continual approach close to a rehearsal approach: we train using only on a subset of the training data, and compare layer parameterizations in this low data regime. 

\subsection{Datasets and pre-trained Models}
\label{sub:data_model}

% introduction of datasets
We conduct our experiments on CIFAR10/CIFAR100 \cite{krizhevskycifar100}  and Core50 \cite{Lomonaco2017core50}. Core50 is a dataset composed of 50 objects from 10 categories (e.g. glasses, phone, cup...), filmed in 11 environments (8 for training and 3 for evaluation). We denote Core10, the 10 classes version where we only predict the category, and Core50, the full 50 classes version. We use both to create various types of scenarios.

The preliminary experiments consist of testing the basic output layer (without masking) to select a learning rate (among $0.1$, $0.01$ and $0.001$ ) and architecture for pre-trained models (among ResNet, VGG16 and GoogleNet available on the torchvision library \cite{NEURIPS2019_9015}). 
While choosing pre-trained models, we made sure that pre-training had not been done using the same dataset that we used in our continual learning experiments (e.g. we did not choose a model pre-trained on CIFAR10 for our CIFAR10 experiments), which would otherwise be "cheating" as the pre-training would have been made using all available data, and not been restricted to data sequentially made available as in continual learning.

%Following the results (appendix \ref{ap:preliminary_experiments}),
To maximize the performance of all layers, we selected hyper-parameters and architecture on i.i.d. settings with all data. We selected a learning rate of $0.01$ for the original linear layer and original linear layer without bias (\textit{Linear\_no\_bias}) as well as for their masked counterpart and $0.1$ for the other layers. Secondly, we found that for Core50 and Core10 experiments, the ResNet model performs best. 
We also eliminated the CIFAR100 dataset from our incremental experiments because the pre-trained model we tested (pre-trained on CIFAR10) did not permit us to achieve a good enough accuracy. We kept CIFAR10, Core10, and Core50 for subsequent experiments. We added CUB200 and a modified version of CIFAR100 (referred to as CIFAR100Lifelong), for continual experiments.
Moreover, we see that 5 epochs per task are sufficient on these datasets, and we will keep this number of epochs for all experiments. 

\subsection{Continual Experiments Settings}
\label{sub:CL_exps}

In these experiments, we evaluate the capacity of every output layer to learn continually with a fixed feature extractor. %. 
We evaluate the CL performance on virtual concept drifts (incremental scenario), on domain drifts (lifelong scenario) (cf Sec.~\ref{sub:linear_cl}). Incremental settings consist of a sequence of tasks where all new tasks bring new unknown classes. Lifelong settings consist of a sequence of tasks where each new task brings new examples of known classes. We also evaluate the best layers in a mixed scenario with both kinds of drifts in the end.
Our scenarios can be reproduced easily with \textit{continuum} library \cite{douillard2021continuum}.
Our experiments use the single-head framework \cite{Farquhar18}, as opposed to multi-head where the task id is used for inference to pre-select a subset of classes.
We do not use task labels during training or testing. The change in labels can signal a change of tasks. However, it does not trigger any continual mechanisms in our experiments, and all training are done using stochastic gradient descent with a momentum of $0.9$.

\begin{itemize}
\item \textbf{Incremental Scenarios: CIFAR10, Core50, CUB200}: Incremental scenarios are characterized by a virtual concept drift between two tasks. These settings evaluate the capacity of learning incremental new classes and distinguishing them from the others. The core50 scenario is composed of 10 tasks; each of them is composed of 5 classes, the CIFAR10 scenario is composed of 5 tasks with 2 classes each, and the CUB200 scenario is composed of 10 tasks of 20 classes each.

\item \textbf{Lifelong Scenarios: Core10Lifelong, CIFAR100Lifelong: } Lifelong settings are characterized by a domain drift between two tasks. The classes stay the same, but the instances change. These settings evaluate the capacity of improving at classifying with new data. The \textbf{Core10Lifelong} scenario is composed of 8 tasks with 10 classes in a given environment, each new task, we visit a new environment. \textbf{CIFAR100Lifelong} is composed of 5 tasks, with 20 labels each that also stay the same. Data are labeled with the coarse labels of CIFAR100. However, data are shared between tasks using the original label to ensure a domain drift between tasks. % (More detail in appendix in section \ref{ap:scenario}).

\item \textbf{Mixed Scenario: Core10Mix: } In this scenario, a new task is triggered by either a virtual concept drift or a domain drift. In this case, the domain drift is characterized by the transition from one object to another in the same category. This scenario is composed of 50 tasks, where each task corresponds to a new object (with category annotation) in all its training domains. So a new task is always a new object, but it is either data from a new class or from a known class visited through another object.
\end{itemize}

The results for those experiments are in Sec.~\ref{sub:continuall_exps}. We report at each epoch the average accuracy over the whole test set of the scenario (with data of all tasks). This measure makes it possible to report the accuracy on a fixed test set and makes the visualization of progress in solving the complete scenario easier.

\subsection{Subset Experiments Settings}

In many continual learning approaches, a subset of the training is saved, either for rehearsal purposes or for validation purposes \cite{buzzega2020dark,douillard2020podnet,prabhu2020gdumb}. In this experiment, we evaluate the capacity of each layer to be trained from scratch with such subsets. It could be more efficient to continually train the feature extractor in a continual end-to-end setting and learn a suitable output layer afterward from a subset of data. This procedure is very fast, and it avoids all problems that projection drifts or any other drift can create. The only (not so simple) problem to solve is which examples to store and how many. Nevertheless, the successes in few-shot learning \cite{Lake11,Fei-Fei06,Wang19} or fast adaptation \cite{caccia2020online} show that when the feature space is good enough, only a small set of data is enough to identify a class.

\section{Results}
\subsection{Continual Experiments}
\label{sub:continuall_exps}

\begin{figure}[h]
    \centering
        \includegraphics[width=\linewidth]{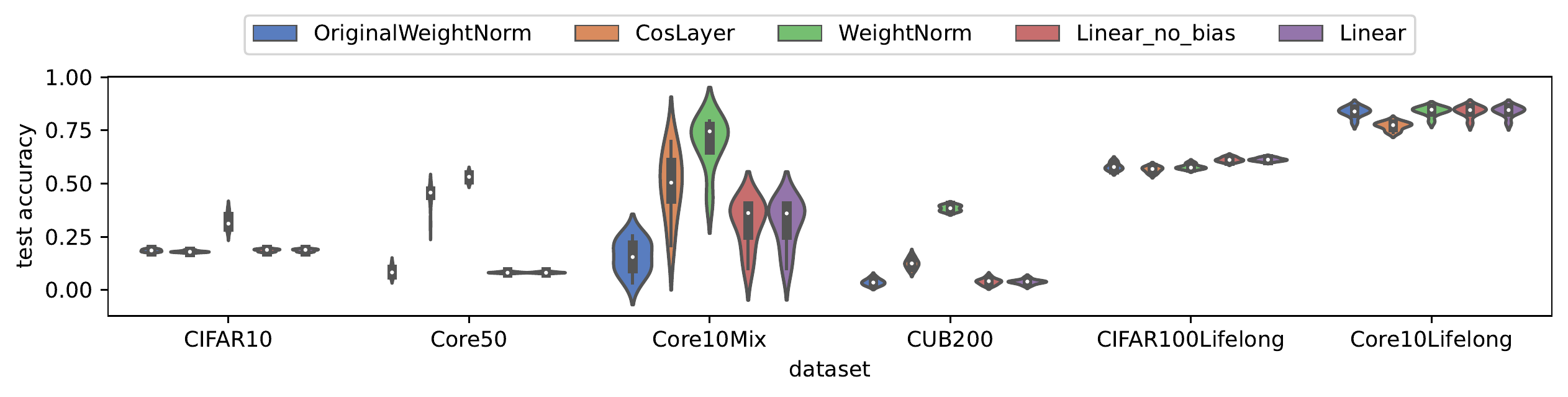}
        \caption{Linear layers comparison: Each layer has been experimented on on 8 different task orders for each scenario.}
        \label{fig:grad_exp}
\end{figure}

\begin{figure*}[h]
    
    \centering
    \begin{subfigure}[t]{0.3\linewidth}
        \centering
        \includegraphics[width=\linewidth]{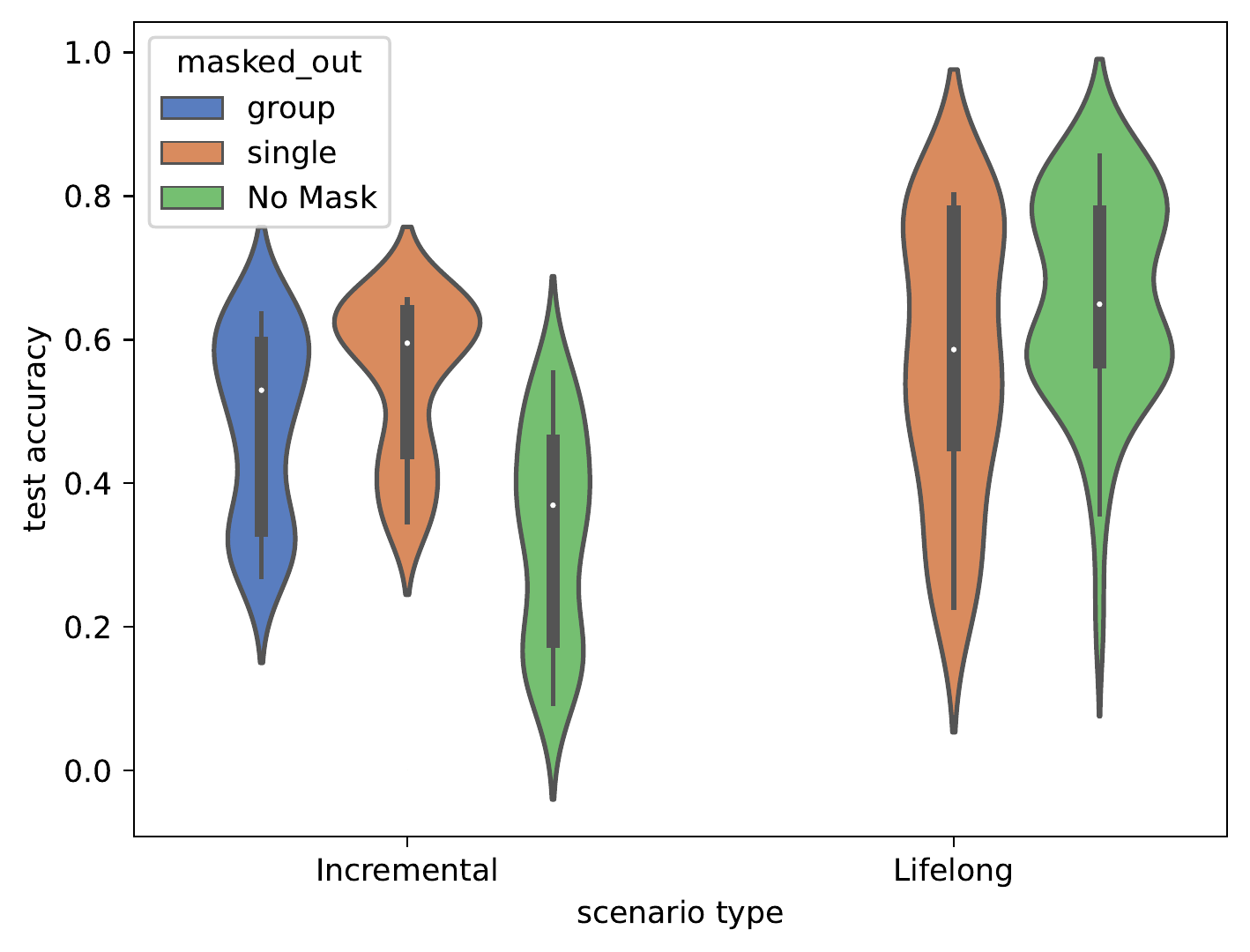}
        \caption{Incremental vs Lifelong}
        \label{fig:mask_inc-lif}
    \end{subfigure}
    \begin{subfigure}[t]{0.3\linewidth}
        \centering
        \includegraphics[width=\linewidth]{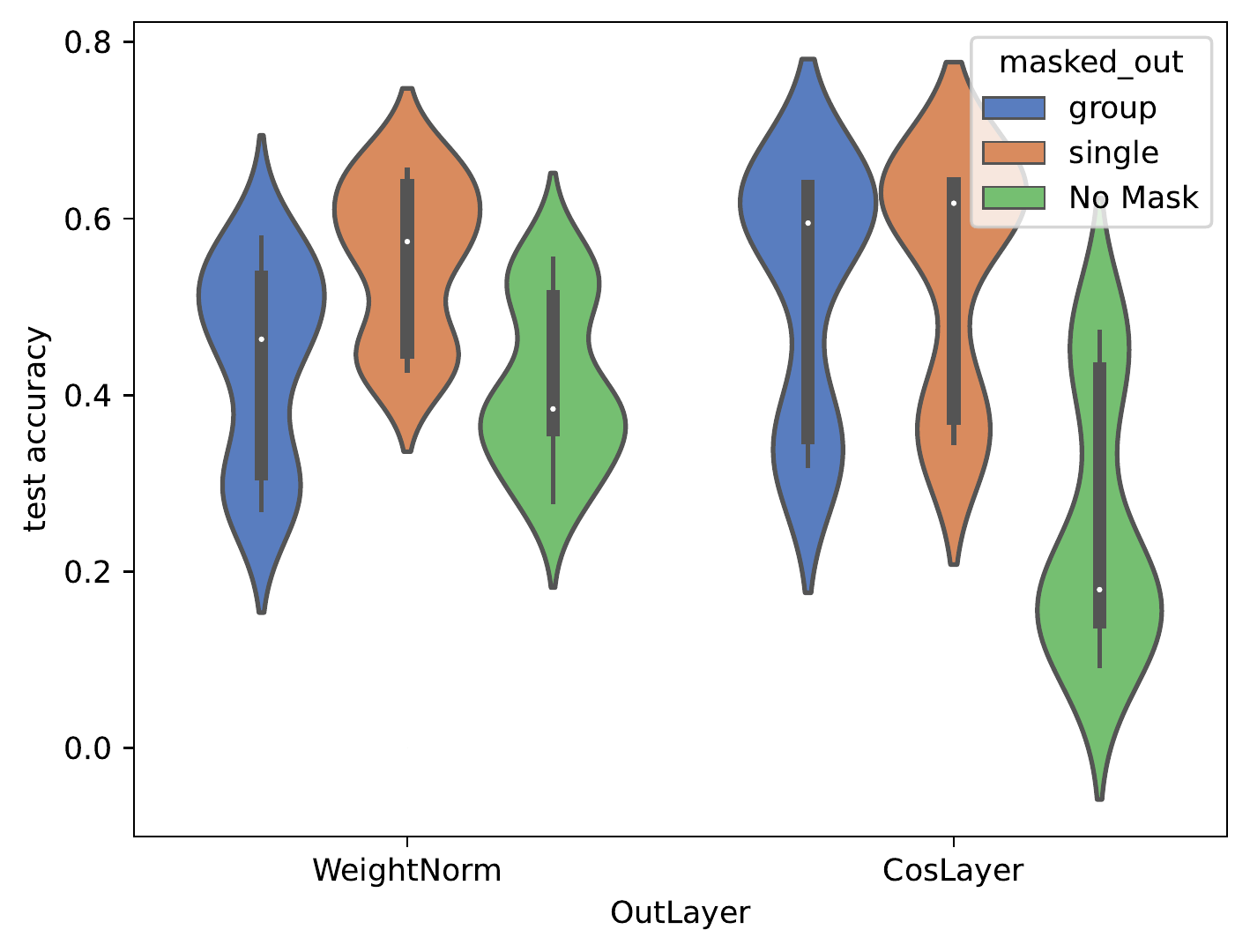}
        \caption{Incremental Scenarios}
        \label{fig:mask_inc}
    \end{subfigure}
    \begin{subfigure}[t]{0.3\linewidth}
        \centering
        \includegraphics[width=\linewidth]{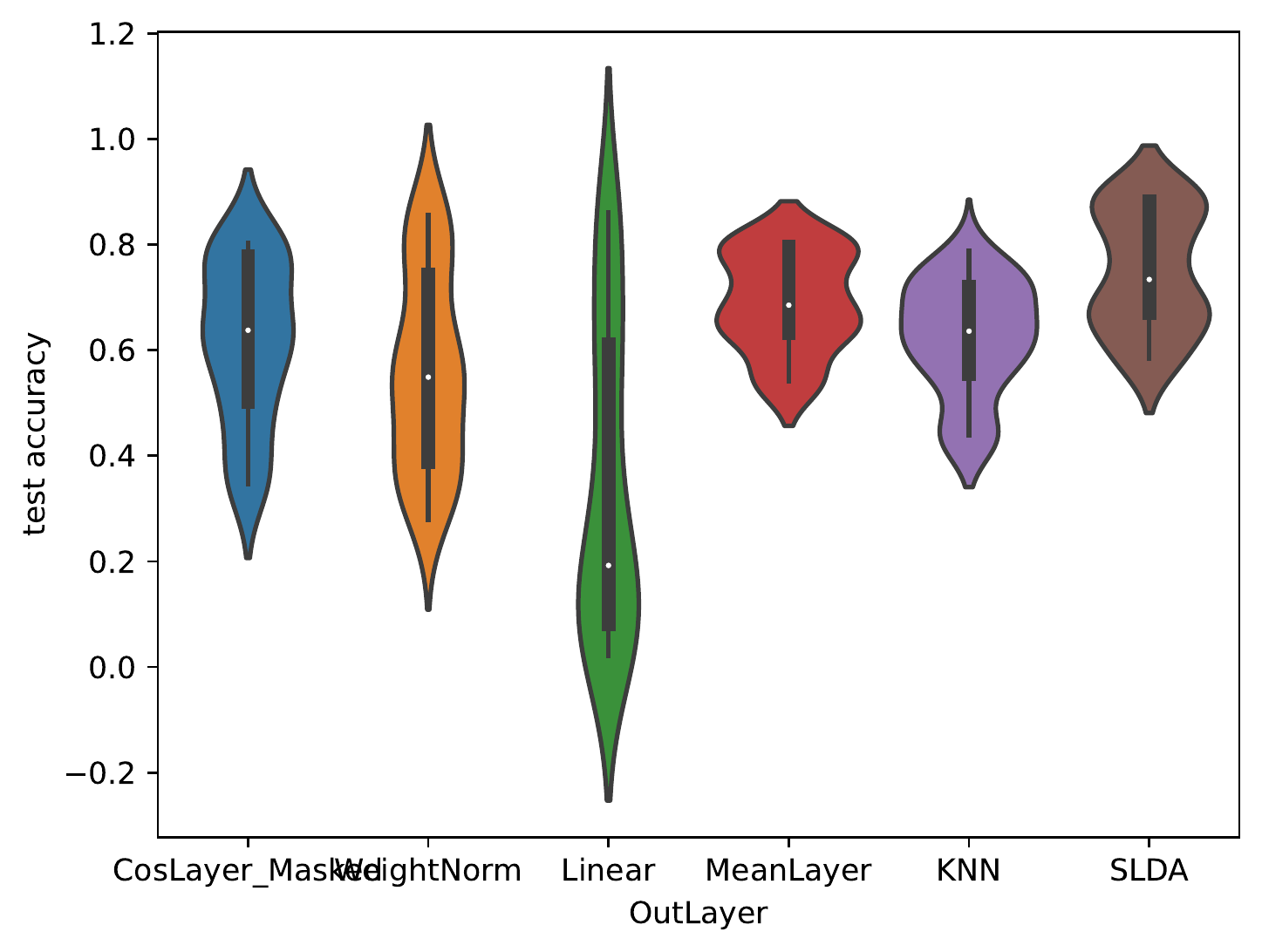}
        \caption{gradient vs no gradient}
        \label{fig:grad_sim}
    \end{subfigure}

    \caption{Experiments in continual scenarios. Highlights of masking effects and similarity based layers.}
    \label{fig:continual_exps}
\end{figure*}

We split the results of continual experiments into four steps; first (Fig.~\ref{fig:grad_exp}),  we compare the various parameterizations of a linear layer presented in Sec.~\ref{sec:layers}, secondly (Fig.~\ref{fig:mask_inc-lif}), we compare the effect of masking in incremental and lifelong scenarios, thirdly (Fig.~\ref{fig:mask_inc}), we analysis which layer (among the selected in the first step) benefits the most from masking, finally (Fig.~\ref{fig:grad_sim}), we compare the best combinations of layer-masking with layers that are not trained by gradient descent, namely: KNN, MeanLayer, and SLDA.  (NB: we keep also the original linear layer as baseline).

The results presented in the various figures lead us to the following conclusions:

\begin{itemize}

\item \textbf{Among the reparameterizations of linear layers, \textit{Coslayer} and \textit{WeightNorm} are the overall best performing.} Indeed, in incremental settings, they are usually outperforming other layers (cf Fig. \ref{fig:grad_exp}).
Both of them only rely on the angle between data and output vectors, which reduces the risk of interference as described in Sec.~\ref{sec:layers}.
The only difference between these layers is that in WeightNorm, the predictions are weighted by the norm of the latent vectors (cf Coslayer eq. \ref{eq:coslayer}, WeightNorm eq. \ref{eq:weightnorm}), which scales the gradients thus may change the learning dynamics. 

\item \textbf{Linear layers do not perform similarly while exposed to different types of drift.} A tendency that emerges from these experiments (Fig. \ref{fig:grad_exp}) is that all layers are compatible with lifelong learning even if masking may create some instability. While, in incremental and mixed scenarios, the layers relying only on angles between data and output vectors and masking helps to avoid weight modification.

\item \textbf{Removing the bias on the linear layer has no effect.} In figure \ref{fig:grad_exp}, we do not observe any improvement by removing the bias from the linear layers in our experiments. This questions \cite{wu2019large}'s observations that correcting the bias can mitigate catastrophic forgetting (CF).  Nevertheless, we hypothesize that the effect of the bias in continual learning depends on the task at hand.

\item \textbf{Masking is efficient in incremental settings to avoid weight modifications.}
The second step (Fig.~\ref{fig:mask_inc-lif}) compares the effect of masking in incremental and lifelong scenarios, by averaging results of all layers in scenarios of both types and comparing them with their masked counterpart. The figure \ref{fig:mask_inc-lif} shows clearly that incremental scenarios benefit the most from masking in particular single masking.
The figure \ref{fig:mask_inc-lif}, compares layers in the incremental scenarios. It shows us that the layer that benefits the most from masking is \textit{Coslayer}, it even outperforms all the other combinations. Hence, for later experiment we keep the single masked coslayer ( \textit{Coslayer\_Masked} ) and the \textit{WeightNorm} as the best not masked layer.

\item \textbf{In lifelong, we cannot avoid weight modification with masking.} In lifelong scenarios, tasks use the same output vectors, i.e., the same weights; 
Hence, masking is less interesting than in incremental scenarios, indeed, the figure \ref{fig:mask_inc-lif} shows some degradation of the original performance in lifelong scenarios.
Note that in lifelong group masking is the same as no masking since all classes are in all tasks.

\item \textbf{Similarity-based layers perform well regardless of the data drift type} (fig.~\ref{fig:grad_sim}).
These layers are easy to train since they need to see the data only once to achieve their best performance. 
In fig.~\ref{fig:grad_exp} these layers over all scenarios perform well in comparison with the best linear layers, especially SLDA.
Nevertheless, those layers can not be used to backpropagate gradient and therefore can not be used to finetune the encoder and necessarily rely on a pre-trained model.
They could, however, be associated with another layer to train the feature extractor as in iCARL \cite{rebuffi2017icarl}.

\end{itemize}

Those experiments show us that the reparametrization of the output layer such as \textit{weightnorm} and \textit{coslayer} improves the ability to learn continually in comparison with the original linear layer. It also indicates the best parametrization depends on the data distribution drifts. In the next section, to better understand how the various layers could perform with a continual algorithm. We study the various layers trained iid with only a subset of the data, it simulates an approach of continual learning with replay.

\subsection{Training with a Subset of Data}

%%%%%%%%%%%%%%%%%%%%%%%%%%%%%%%%%%%%%%%%%%%%%%%%%%%%%%%%%%%%%%%%%%%
\begin{table*}
\centering

  \caption{Subset experiments: Mean Accuracy and standard deviation on 8 runs with different seeds on Core50. In bold, best performing layers, in the subset setting.}
  \label{tab:subsets_results}
 
  \begin{tabular}{ccccccc}
    \hline 
    OutLayer&Dataset $\backslash$ Subset & 100 & 200 & 500 & 1000 & All\\ 
    \hline

    % ##############  0.1 ############## 
CosLayer & 
Core50 & 
$10.30$\mbox{\scriptsize{$\pm1.98$}}& % 100  
$12.43$\mbox{\scriptsize{$\pm3.56$}}& % 200  
$18.11$\mbox{\scriptsize{$\pm4.17$}}& % 500  
$26.48$\mbox{\scriptsize{$\pm4.16$}}& % 1000  
$53.93$\mbox{\scriptsize{$\pm0.90$}}\\ 
 
\rowcolor{gray!20}\textbf{WeightNorm} & 
Core50 & 
$32.06$\mbox{\scriptsize{$\pm2.93$}}& % 100  
$44.32$\mbox{\scriptsize{$\pm1.92$}}& % 200  
$61.91$\mbox{\scriptsize{$\pm1.10$}}& % 500  
$69.44$\mbox{\scriptsize{$\pm0.67$}}& % 1000  
$77.04$\mbox{\scriptsize{$\pm0.33$}}\\ 
 
\textbf{OriginalWeightNorm} & 
Core50 & 
\textbf{$33.19$}\mbox{\scriptsize{$\pm2.62$}}& % 100  
$45.68$\mbox{\scriptsize{$\pm1.96$}}& % 200  
$62.39$\mbox{\scriptsize{$\pm1.01$}}& % 500  
$69.17$\mbox{\scriptsize{$\pm0.65$}}& % 1000  
$76.85$\mbox{\scriptsize{$\pm0.61$}}\\ 
 
\rowcolor{gray!20}CosLayer-Masked & 
Core50 & 
$30.62$\mbox{\scriptsize{$\pm2.57$}}& % 100  
$41.22$\mbox{\scriptsize{$\pm1.68$}}& % 200  
$56.47$\mbox{\scriptsize{$\pm1.13$}}& % 500  
$63.10$\mbox{\scriptsize{$\pm0.62$}}& % 1000  
$63.43$\mbox{\scriptsize{$\pm5.78$}}\\ 
 
WeightNorm-Masked & 
Core50 & 
$12.95$\mbox{\scriptsize{$\pm2.07$}}& % 100  
$14.98$\mbox{\scriptsize{$\pm2.89$}}& % 200  
$21.13$\mbox{\scriptsize{$\pm1.22$}}& % 500  
$23.42$\mbox{\scriptsize{$\pm2.49$}}& % 1000  
$31.18$\mbox{\scriptsize{$\pm17.50$}}\\ 
 
\rowcolor{gray!20}OriginalWeightNorm-Masked & 
Core50 & 
$12.27$\mbox{\scriptsize{$\pm2.59$}}& % 100  
$17.42$\mbox{\scriptsize{$\pm2.52$}}& % 200  
$20.82$\mbox{\scriptsize{$\pm4.94$}}& % 500  
$22.21$\mbox{\scriptsize{$\pm5.50$}}& % 1000  
$22.43$\mbox{\scriptsize{$\pm4.36$}}\\ 
 
\textbf{Linear} & 
Core50 & 
$32.80$\mbox{\scriptsize{$\pm2.67$}}& % 100  
$44.88$\mbox{\scriptsize{$\pm1.94$}}& % 200  
$61.80$\mbox{\scriptsize{$\pm1.06$}}& % 500  
$69.03$\mbox{\scriptsize{$\pm0.65$}}& % 1000  
$76.29$\mbox{\scriptsize{$\pm0.34$}}\\ 
 
% \rowcolor{gray!20}Linear-no-bias & 
% Core50 & 
% $32.80$\mbox{\scriptsize{$\pm2.68$}}& % 100  
% $44.89$\mbox{\scriptsize{$\pm1.94$}}& % 200  
% $61.80$\mbox{\scriptsize{$\pm1.06$}}& % 500  
% $69.02$\mbox{\scriptsize{$\pm0.65$}}& % 1000  
% $76.29$\mbox{\scriptsize{$\pm0.34$}}\\ 
 
\rowcolor{gray!20} Linear-Masked & 
Core50 & 
$7.37$\mbox{\scriptsize{$\pm1.98$}}& % 100  
$8.20$\mbox{\scriptsize{$\pm2.68$}}& % 200  
$14.02$\mbox{\scriptsize{$\pm5.88$}}& % 500  
$18.46$\mbox{\scriptsize{$\pm2.60$}}& % 1000  
$27.25$\mbox{\scriptsize{$\pm28.39$}}\\ 
 
Linear-no-bias-Masked & 
Core50 & 
$7.36$\mbox{\scriptsize{$\pm1.97$}}& % 100  
$8.22$\mbox{\scriptsize{$\pm2.67$}}& % 200  
$14.04$\mbox{\scriptsize{$\pm5.88$}}& % 500  
$18.48$\mbox{\scriptsize{$\pm2.55$}}& % 1000  
$25.81$\mbox{\scriptsize{$\pm29.13$}}\\ 
 
\rowcolor{gray!20} KNN & 
Core50 & 
$25.95$\mbox{\scriptsize{$\pm3.09$}}& % 100  
$34.17$\mbox{\scriptsize{$\pm2.14$}}& % 200  
$45.50$\mbox{\scriptsize{$\pm2.04$}}& % 500  
$52.13$\mbox{\scriptsize{$\pm1.19$}}& % 1000  
$65.50$\mbox{\scriptsize{$\pm0.14$}}\\ 
 
SLDA & 
Core50 & 
$17.30$\mbox{\scriptsize{$\pm1.23$}}& % 100  
$30.23$\mbox{\scriptsize{$\pm5.58$}}& % 200  
$17.71$\mbox{\scriptsize{$\pm1.47$}}& % 500  
$60.14$\mbox{\scriptsize{$\pm0.93$}}& % 1000  
$78.55$\mbox{\scriptsize{$\pm0.03$}}\\ 
 
\rowcolor{gray!20} MeanLayer & 
Core50 & 
$28.81$\mbox{\scriptsize{$\pm2.68$}}& % 100  
$40.71$\mbox{\scriptsize{$\pm1.65$}}& % 200  
$56.52$\mbox{\scriptsize{$\pm1.29$}}& % 500  
$63.20$\mbox{\scriptsize{$\pm0.79$}}& % 1000  
$71.51$\mbox{\scriptsize{$\pm0.00$}}\\ 
 
MedianLayer & 
Core50 & 
$26.83$\mbox{\scriptsize{$\pm2.26$}}& % 100  
$36.03$\mbox{\scriptsize{$\pm1.65$}}& % 200  
$53.07$\mbox{\scriptsize{$\pm1.15$}}& % 500  
$60.73$\mbox{\scriptsize{$\pm0.77$}}& % 1000  
$70.22$\mbox{\scriptsize{$\pm0.00$}}\\

 \hline 

    \end{tabular}

\end{table*}
%%%%%%%%%%%%%%%%%%%%%%%%%%%%%%%%%%%%%%%%%%%%%%%%%%%%%%%%%%%%%%%%%%%

\begin{figure}[!h]

    \begin{subfigure}[t]{\linewidth}
        \centering
        \includegraphics[width=\linewidth]{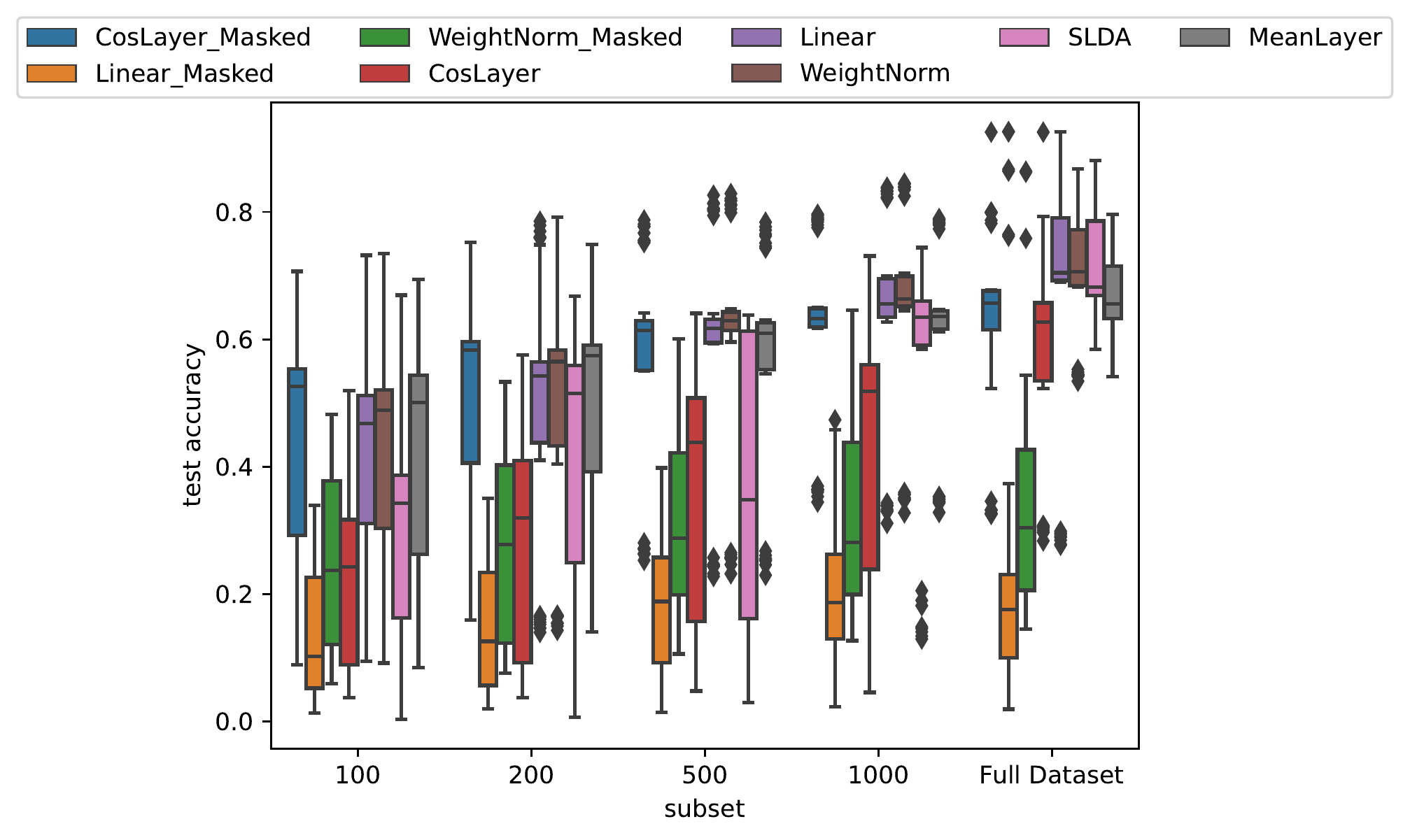}
        \caption{Comparison of performance of a significant set of output layers.}
    \label{fig:subset_layers}
    \end{subfigure}
    
        \centering
    \begin{subfigure}[t]{0.5\linewidth}
        \centering
        \includegraphics[width=\linewidth]{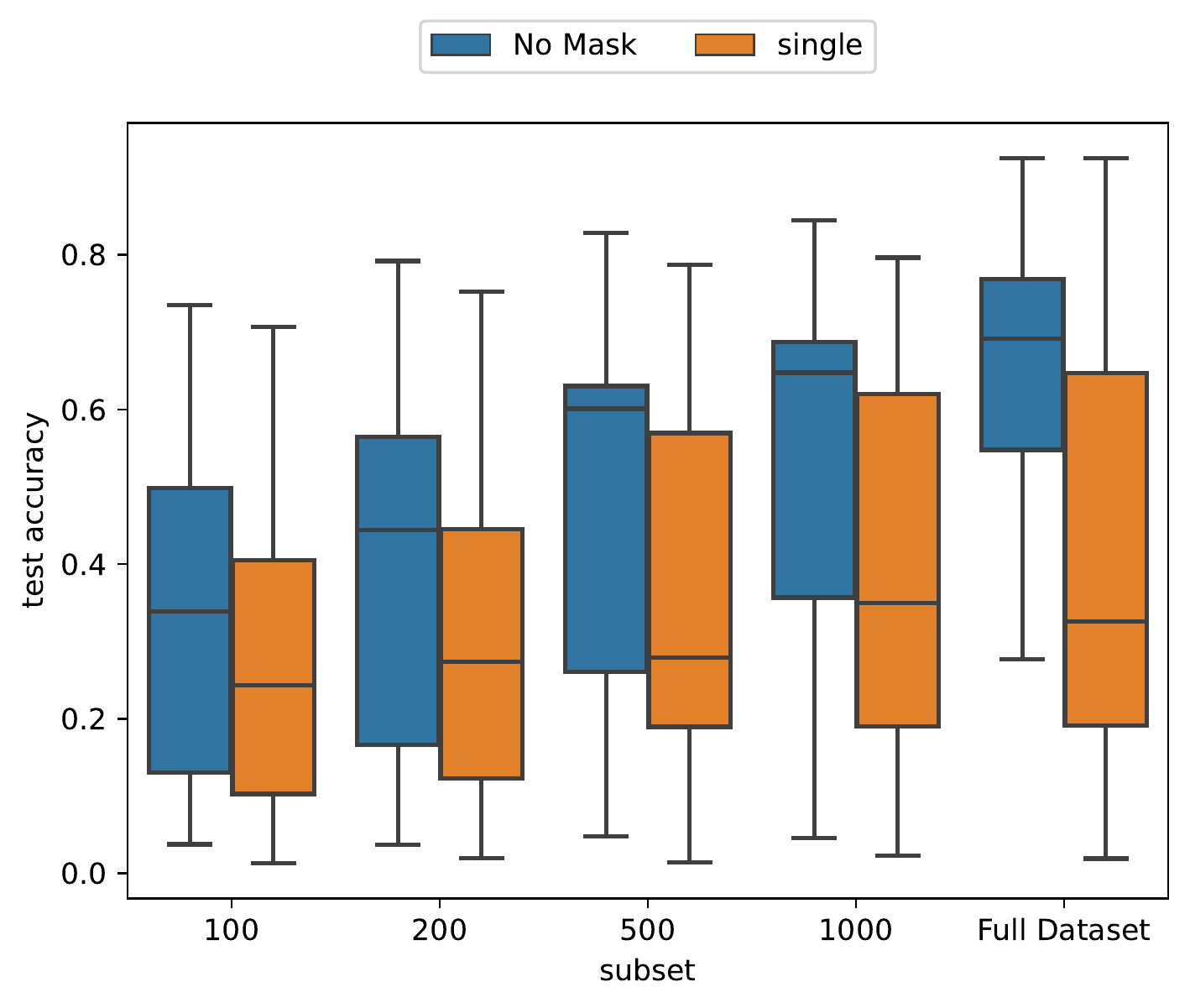}
        \caption{Impact of masking.}
        \label{fig:subset_masked_exp}
    \end{subfigure}

    \caption{Subset Experiments realized on 8 seeds. The test accuracy is on the full test set. Results are averaged over all datasets experimented (CIFAR10, Core50, CUB200, Core10, CIFAR100Lifelong).
    We compare the different layer types in various subset setting. \textit{\_Masked} denotes layer trained with single masking.}
    \label{fig:subset_exp}
\end{figure}

In these experiments, we randomly select a sub-sample of the training dataset, and we train the output layers on it. The experiment's results are gathered in Fig. \ref{fig:subset_exp}.  We show also a representative subset of results with layer performances on Core50 in Table \ref{tab:subsets_results}.

The goal is to measure how the various layers can learn with a low amount of data. It gives us also a good continual baseline close to a replay approach such as GDump \cite{prabhu2020gdumb}. We save samples in the memory, and we train at the end of the scenario only on the memory's samples. This approach performs particularly well in comparison to many continual approaches. It gives us some insight into how layers can perform in few-shot scenarios and replay scenarios with limited data.
Here are some conclusions of these experiments:

%\begin{itemize}

\textbf{ $\bullet$ Masking does not work in an iid setting.} In table \ref{fig:subset_masked_exp}, most of the masked layers underperform, except the CosLayer, which performs better with a mask  (cf Fig. \ref{fig:subset_layers}).

\textbf{$\bullet$ \textit{WeightNorm} and \textit{CosLayer\_Masked} are still very competitive among linear layers.}
In figure  \ref{fig:subset_layers}, they are among the best linear layers. The original linear layer, which was bad performing in continual settings is well-performing when the distribution is i.i.d. Hence, the original linear layer is still a good option for few-shot learning or with replay.

\textbf{$\bullet$ Similarity-based layers tend to slightly under-perform linear layers.} In Fig.~\ref{fig:subset_layers}, they do not perform as well as the best linear layers, losing the advantages they had in continual settings. They can still have a decent accuracy.

%\end{itemize}

These experiments with iid training on a low data regime show that in such settings, the reparameterizations of the linear layer are well-performing. In conclusion of all the experiments, we recommend using WeigthNorm (especially in lifelong) or \textit{CosLayer\_Masked} in continual scenarios whatever the setting and the training procedure.

\section{Application for End-to-End with Replay}
\label{sec:e2e}
\begin{figure}[!h]
        \centering
        \includegraphics[width=0.6\linewidth]{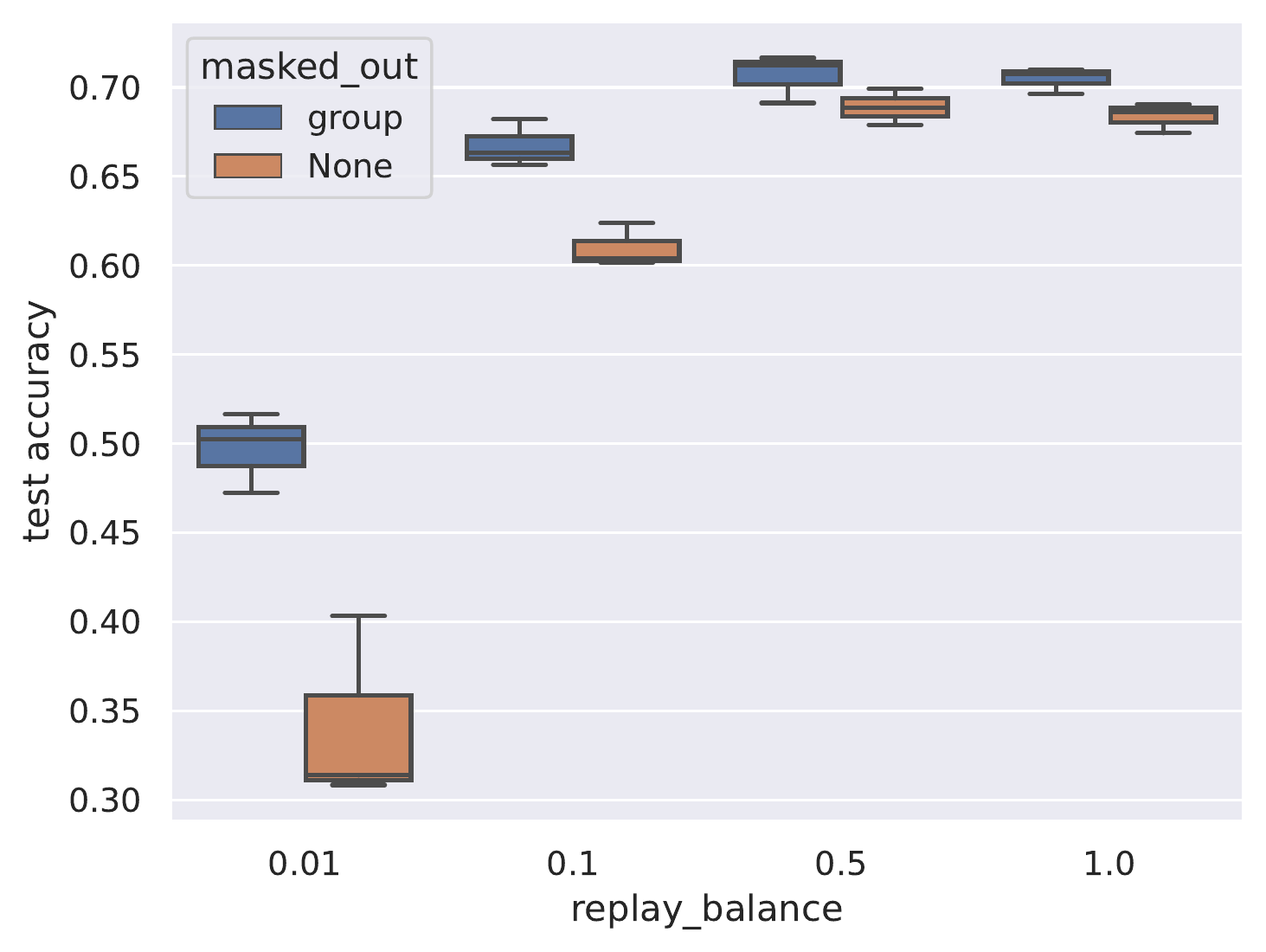}
        \caption{\textbf{Final test accuracy vs. replay balance}: Group masking reduces the need for replay. The replay balance is ratio between instances of a replayed class and a new class.}
        \label{fig:masking_e2e}
\end{figure}
The adaptation of the output layer can also be helpful for end-to-end training. In this example experiment, we show that group masking improves performance with replay, particularly when the amount of replay is low.

 We train a resnet model on a 5-tasks class-incremental setting with CIFAR10. We use the WeightNorm layer with group masking and without masking. We select this baseline since it is the most consistently well performing in our previous experiments. The CL approach is a vanilla replay method with a random selection of samples to populate the buffer (2000 examples per class). We use a large buffer since memory is generally not the main bottleneck. In the experiment, we variate the amount of replay balance to control the replay balance. The replay balance is the ratio between new and old classes' instances. $0.5$ replay balance means two times more instances for new classes.
The lower the replay balance, the more imbalanced the class distribution is, and the lower the total compute is. Indeed, by replaying less, they are fewer examples to process. 
We use a batch size of 8 with 3 seeds. The results are provided in Fig.~\ref{fig:masking_e2e}. Group masking improves final performance, especially with low replay balance. This shows that group masking makes training more compute efficient when using replay and lowers the need for replay.
\section{Discussion}

% 1. conclusion de le section precedente: on adresse wight modificication et  inter-tache  interference dans la derniere couche sous la l'hypothese d'un feature exctractor frozen en evitant le projection drift.
In this paper, we studied and proposed various approaches that mitigate CF in the output layer by addressing weights modification and interference, by restricting ourselves to a setup where there is no projection drift.

% 2. en multi head c'est l'inverse, ils utilisent le task label pour eviter d'avoir du wight modificication et inter-tache interference, et addresse seulement le projection dirft.

On the opposite, multi-head architectures use the test task label and use a head specific to the current task. It avoids interference and weight modification by training separate heads. 
%Indeed, in multi-head the task label serves to select the correct head: classes that belong to another task cannot be predicted. 
Interestingly, in multi-heads settings, using regularization %\cite{kirkpatrick2017overcoming,Zenke17,Ritter18Online}
or dynamic architecture
%\cite{Rusu16progressive,Mallya18Piggyback,Rajasegaran2019Random}
, have been shown to perform well on end-to-end tasks. Then, they are able to deal with projection drifts: the head of task 0 still works when applied to the feature extractor trained up until the last task.

% 3. discuter qu ést ce qu'il manque pour le conintual end-to end : les taches sont tres differentes
% So, if in this paper, we show that with a proper output layer we can deal with interference and weight modification, and with multi-head approaches, we can deal with projection drift. The single head setting could be solved without replay. 

Unfortunately, end-to-end approaches that work in multi-heads can not be directly connected with our findings to create a good end-to-end single head approach. Indeed, single-head models need discriminative inter-task representations that are not necessary for multi-head settings. Therefore, the constraints on the feature extractor are not the same, and direct transfer is not possible.

%
%
% 4. ccl discussion sur conpenser le drift (sans lien à single / multi head)
Dealing with projection drift in single-head settings stays one of the biggest challenges in continual learning. 
Replay might be a simple solution to simulate an iid training and avoid problems with projection drift. However, even if replay might be necessary for continual learning scenarios \cite{diethe2018continual,lesort2019regularization}
, replaying data may have consequences in the memory needs and compute needs. Therefore, it is worth finding training procedures, architectures, or regularizations that do not necessarily replace replay but at least reduce its need as group masking does in experiment of Sec.~\ref{sec:e2e}.

\section{Conclusion}
\label{sec:ccl}

In this paper, we conduct an empirical evaluation of the various output layers of deep neural networks in CL scenarios.
%This evaluation is the first to be conducted where output layers are evaluated independently from feature extractor training.
This evaluation gives us clear insights into how output layers learn continually or on a low data regime. We introduced \textit{WeigthNorm} and \textit{CosLayer\_Masked} layers, that showed clear advantages in all setting types.
We also showed how distribution drifts might affect the output layers differently, namely in lifelong and incremental settings. This perspective should incite researchers to study more the connection between distribution drifts and catastrophic forgetting.

% reparametrization for the win
On one hand, classifiers such as KNN, MeanLayer, and SLDA provide very strong baselines that are very effective with a frozen feature extractor.
On the other hand, our results show that by a reparameterization of the classical linear layer, we can greatly improve performance, especially in incremental settings. This reparameterization helps to reduce the risks of interference and weight modifications. Some output layer types can learn continually when there is no projection drift with stochastic gradient training. These findings could be integrated into all end-to-end training to make continual training more efficient.

\bibliographystyle{abbrv}
\bibliography{main/continual,main/others}

\begin{thebibliography}{10}

\bibitem{ALIYARIGHASSABEH20151999}
Y.~{Aliyari Ghassabeh}, F.~Rudzicz, and H.~A. Moghaddam.
\newblock Fast incremental lda feature extraction.
\newblock {\em Pattern Recognition}, 48(6):1999--2012, 2015.

\bibitem{buzzega2020dark}
P.~Buzzega, M.~Boschini, A.~Porrello, D.~Abati, and S.~Calderara.
\newblock Dark experience for general continual learning: a strong, simple
  baseline.
\newblock In H.~Larochelle, M.~Ranzato, R.~Hadsell, M.~F. Balcan, and H.~Lin,
  editors, {\em Advances in Neural Information Processing Systems 33}, pages
  15920--15930. Curran Associates, Inc., 2020.

\bibitem{caccia2021reducing}
L.~Caccia, R.~Aljundi, T.~Tuytelaars, J.~Pineau, and E.~Belilovsky.
\newblock Reducing representation drift in online continual learning.
\newblock {\em arXiv preprint arXiv:2104.05025}, 2021.

\bibitem{caccia2020online}
M.~Caccia, P.~Rodriguez, O.~Ostapenko, F.~Normandin, M.~Lin, L.~Caccia,
  I.~Laradji, I.~Rish, A.~Lacoste, D.~Vazquez, and L.~Charlin.
\newblock Online fast adaptation and knowledge accumulation: a new approach to
  continual learning.
\newblock {\em NeurIPS}, 2020.

\bibitem{Chatterjee1997self}
C.~Chatterjee and V.~Roychowdhury.
\newblock On self-organizing algorithms and networks for class-separability
  features.
\newblock {\em IEEE Transactions on Neural Networks}, 8(3):663--678, 1997.

\bibitem{DEMIR2005421}
G.~Demir and K.~Ozmehmet.
\newblock Online local learning algorithms for linear discriminant analysis.
\newblock {\em Pattern Recognition Letters}, 26(4):421--431, 2005.
\newblock ICAPR 2003.

\bibitem{diethe2018continual}
T.~Diethe, T.~Borchert, E.~Thereska, B.~d.~B. Pigem, and N.~Lawrence.
\newblock Continual learning in practice.
\newblock In {\em NeurIPS Continual Learning Workshop}, 2018.

\bibitem{douillard2020podnet}
A.~Douillard, M.~Cord, C.~Ollion, T.~Robert, and E.~Valle.
\newblock Podnet: Pooled outputs distillation for small-tasks incremental
  learning.
\newblock In {\em Proceedings of the IEEE European Conference on Computer
  Vision (ECCV)}, 2020.

\bibitem{douillard2021continuum}
A.~Douillard and T.~Lesort.
\newblock Continuum: Simple management of complex continual learning scenarios,
  2021.

\bibitem{Farquhar18}
S.~Farquhar and Y.~Gal.
\newblock Towards robust evaluations of continual learning.
\newblock {\em arXiv preprint arXiv:1805.09733}, 2018.

\bibitem{Fei-Fei06}
L.~Fei-Fei, R.~Fergus, and P.~Perona.
\newblock One-shot learning of object categories.
\newblock {\em IEEE transactions on pattern analysis and machine intelligence},
  28(4):594--611, 2006.

\bibitem{hayes2020lifelong}
T.~L. Hayes and C.~Kanan.
\newblock Lifelong machine learning with deep streaming linear discriminant
  analysis.
\newblock In {\em Proceedings of the IEEE/CVF Conference on Computer Vision and
  Pattern Recognition Workshops}, pages 220--221, 2020.

\bibitem{he2016deep}
K.~He, X.~Zhang, S.~Ren, and J.~Sun.
\newblock Deep residual learning for image recognition.
\newblock In {\em Proceedings of the IEEE conference on computer vision and
  pattern recognition}, pages 770--778, 2016.

\bibitem{Hou_2019_CVPR}
S.~Hou, X.~Pan, C.~C. Loy, Z.~Wang, and D.~Lin.
\newblock Learning a unified classifier incrementally via rebalancing.
\newblock In {\em The IEEE Conference on Computer Vision and Pattern
  Recognition (CVPR)}, June 2019.

\bibitem{ioffe2015batch}
S.~Ioffe and C.~Szegedy.
\newblock Batch normalization: Accelerating deep network training by reducing
  internal covariate shift.
\newblock In {\em International conference on machine learning}, pages
  448--456. PMLR, 2015.

\bibitem{kirkpatrick2017overcoming}
J.~Kirkpatrick, R.~Pascanu, N.~Rabinowitz, J.~Veness, G.~Desjardins, A.~A.
  Rusu, K.~Milan, J.~Quan, T.~Ramalho, A.~Grabska-Barwinska, et~al.
\newblock Overcoming catastrophic forgetting in neural networks.
\newblock {\em Proc. of the national academy of sciences}, 2017.

\bibitem{krizhevskycifar100}
A.~Krizhevsky and G.~Hinton.
\newblock Learning multiple layers of features from tiny images.
\newblock {\em Technical Report}, 2009.

\bibitem{Lake11}
B.~Lake, R.~Salakhutdinov, J.~Gross, and J.~Tenenbaum.
\newblock One shot learning of simple visual concepts.
\newblock In {\em Proceedings of the Annual Meeting of the Cognitive Science
  Society}, volume~33, 2011.

\bibitem{lesort2021understanding}
T.~Lesort, M.~Caccia, and I.~Rish.
\newblock Understanding continual learning settings with data distribution
  drift analysis.
\newblock {\em arXiv preprint arXiv:2104.01678}, 2021.

\bibitem{Lesort18}
T.~Lesort, N.~D{\'{\i}}az-Rodr{\'{\i}}guez, J.-F. Goudou, and D.~Filliat.
\newblock State representation learning for control: An overview.
\newblock {\em Neural Networks}, 2018.

\bibitem{lesort2019regularization}
T.~Lesort, A.~Stoian, and D.~Filliat.
\newblock Regularization shortcomings for continual learning.
\newblock {\em arXiv preprint arXiv:1912.03049}, 2019.

\bibitem{Lomonaco2017core50}
V.~Lomonaco and D.~Maltoni.
\newblock Core50: a new dataset and benchmark for continuous object
  recognition.
\newblock In S.~Levine, V.~Vanhoucke, and K.~Goldberg, editors, {\em
  Proceedings of the 1st Annual Conference on Robot Learning}, volume~78 of
  {\em Proceedings of Machine Learning Research}, pages 17--26. PMLR, 13--15
  Nov 2017.

\bibitem{lomonaco2020rehearsal}
V.~Lomonaco, D.~Maltoni, and L.~Pellegrini.
\newblock Rehearsal-free continual learning over small non-iid batches.
\newblock In {\em 2020 IEEE/CVF Conference on Computer Vision and Pattern
  Recognition Workshops (CVPRW)}, pages 989--998. IEEE Computer Society, 2020.

\bibitem{NEURIPS2019_9015}
A.~Paszke, S.~Gross, F.~Massa, A.~Lerer, J.~Bradbury, G.~Chanan, T.~Killeen,
  Z.~Lin, N.~Gimelshein, L.~Antiga, A.~Desmaison, A.~Kopf, E.~Yang, Z.~DeVito,
  M.~Raison, A.~Tejani, S.~Chilamkurthy, B.~Steiner, L.~Fang, J.~Bai, and
  S.~Chintala.
\newblock Pytorch: An imperative style, high-performance deep learning library.
\newblock In H.~Wallach, H.~Larochelle, A.~Beygelzimer, F.~dAlch\'{e} Buc,
  E.~Fox, and R.~Garnett, editors, {\em Advances in Neural Information
  Processing Systems 32}, pages 8024--8035. Curran Associates, Inc., 2019.

\bibitem{prabhu2020gdumb}
A.~Prabhu, P.~H. Torr, and P.~K. Dokania.
\newblock Gdumb: A simple approach that questions our progress in continual
  learning.
\newblock In {\em European Conference on Computer Vision}, pages 524--540.
  Springer, 2020.

\bibitem{raffin2019decoupling}
A.~Raffin, A.~Hill, K.~R. Traor{\'e}, T.~Lesort, N.~D{\'\i}az-Rodr{\'\i}guez,
  and D.~Filliat.
\newblock Decoupling feature extraction from policy learning: assessing
  benefits of state representation learning in goal based robotics.
\newblock {\em Workshop on “Structure and Priors in Reinforcement Learning”
  (SPiRL) at ICLR}, 2019.

\bibitem{rebuffi2017icarl}
S.-A. Rebuffi, A.~Kolesnikov, G.~Sperl, and C.~H. Lampert.
\newblock icarl: Incremental classifier and representation learning.
\newblock In {\em Proceedings of the IEEE Conference on Computer Vision and
  Pattern Recognition}, pages 2001--2010, 2017.

\bibitem{salimans2016weight}
T.~Salimans and D.~P. Kingma.
\newblock Weight normalization: A simple reparameterization to accelerate
  training of deep neural networks.
\newblock {\em arXiv preprint arXiv:1602.07868}, 2016.

\bibitem{PangIncremental2005}
{Shaoning Pang}, S.~{Ozawa}, and N.~{Kasabov}.
\newblock Incremental linear discriminant analysis for classification of data
  streams.
\newblock {\em IEEE Transactions on Systems, Man, and Cybernetics, Part B
  (Cybernetics)}, 35(5):905--914, 2005.

\bibitem{simonyan2014very}
K.~Simonyan and A.~Zisserman.
\newblock Very deep convolutional networks for large-scale image recognition.
\newblock {\em arXiv preprint arXiv:1409.1556}, 2014.

\bibitem{stooke2020decoupling}
A.~Stooke, K.~Lee, P.~Abbeel, and M.~Laskin.
\newblock Decoupling representation learning from reinforcement learning.
\newblock {\em arXiv preprint arXiv:2009.08319}, 2020.

\bibitem{szegedy2014going}
C.~Szegedy, W.~Liu, Y.~Jia, P.~Sermanet, S.~Reed, D.~Anguelov, D.~Erhan,
  V.~Vanhoucke, and A.~Rabinovich.
\newblock Going deeper with convolutions, 2014.

\bibitem{van2019three}
G.~M. van~de Ven and A.~S. Tolias.
\newblock Three scenarios for continual learning.
\newblock {\em arXiv preprint arXiv:1904.07734}, 2019.

\bibitem{Wang19}
W.~Wang, V.~W. Zheng, H.~Yu, and C.~Miao.
\newblock A survey of zero-shot learning: Settings, methods, and applications.
\newblock {\em ACM Trans. Intell. Syst. Technol.}, 10(2):13:1--13:37, Jan.
  2019.

\bibitem{wu2019large}
Y.~Wu, Y.~Chen, L.~Wang, Y.~Ye, Z.~Liu, Y.~Guo, and Y.~Fu.
\newblock Large scale incremental learning.
\newblock In {\em Proceedings of the IEEE/CVF Conference on Computer Vision and
  Pattern Recognition}, pages 374--382, 2019.

\bibitem{zhao2020maintaining}
B.~Zhao, X.~Xiao, G.~Gan, B.~Zhang, and S.-T. Xia.
\newblock Maintaining discrimination and fairness in class incremental
  learning.
\newblock In {\em Proceedings of the IEEE/CVF Conference on Computer Vision and
  Pattern Recognition}, pages 13208--13217, 2020.

\end{thebibliography}

\newpage
\appendix

\section*{Appendix}

\section{Scenarios description}
\label{ap:scenario}

\subsection{CIFAR10 5 tasks}
A classical incremental setting, with 2 classes per task with original CIFAR10 dataset.

\subsubsection{CIFAR100Lifelong 5 tasks}
CIFAR100 dataset has two labelization systems: the classes labels and the coarse labels. There are 20 coarse labels gathering 5 classes each. 
To create CIFAR100Lifelong 5 tasks, we share data into tasks such as having one data for each coarse label in each task. 
On the other hand, we use class labels to ensure that the data is different for each task.

We have then 5 tasks with data for all coarse labels and from 20 different classes.
A class does not appear in 2 tasks, as in an incremental. However, the model should predict the coarse label.
We obtain a scenario with 5 tasks with always the same 20 labels. However, the data change, and the model should learn continually to predict the coarse labels for data for all classes correctly.  This scenario makes it possible to measure the resistance to mode change (change of class) while continually learning the coarse label classification.

\subsection{Core50 10 tasks}

A classical incremental setting, with 5 classes per task with the 50 objects core50.

\subsection{Core10Lifelong 8 tasks}

Core50 is composed of 50 objects, equitably shared into 10 classes filmed in 11 environments (8 environments for train and 3 for tests.) In Core10Lifelong, we have 8 tasks where we explore each time one environment with all objects. Objects are annotated with their class labels. Hence, we have 8 tasks of 10-way classification. The object stays the same from one task to another; only the environment changes. This scenario makes it possible to measure the resistance to background change while learning continually. 

\subsection{Core10Mix 50 tasks}

In this scenario, each task contains the data corresponding to one object in all of its environments. We have hence a sequence of 50 tasks. However, objects are annotated with their classes labels. Hence all classes are revisited 5 times in the scenario (in a random way). This scenario is a mixture of incremental and lifelong to measure the ability to solve a scenario with several types of data distribution drifts.

\section{SLDA Layer}
\label{ap:slda}

The approach we use (proposed by \cite{hayes2020lifelong}) stores one mean vector per class $\mu_k$ with a counter $c_k \in \mathbb{R}$ as well as a covariance matrix of the same size as $A$ (common to all classes).
The covariance matrix's update is done with:
\begin{equation}
 \Sigma_{t+1} = \frac{t \Sigma_t + \Delta_t}{t+1}
\label{eq:covariance_slda}
\end{equation}
Where $\Delta_t$ is:
\begin{equation}
 \Delta_t = \frac{t(z_t - \mu_{k})^2}{t+1}
\label{eq:delta_slda}
\end{equation}
For prediction, SLDA uses a precision matrix $\Lambda=[(1-\epsilon)\Sigma + \epsilon I]^{-1}$, where $\epsilon=10^{-4}$ is called the shrinkage parameter and $I \in \mathbb{R}^{d\times d}$ is the identity matrix.

Then two vectors $w$ and $b$ are computed
$w_k = \Lambda \mu_k$, 
$b_k= - \frac{1}{2} (\mu_k^\top  \Lambda \mu_k)$

And finally:
$o_k =  \langle z_t, w_k  \rangle  + b_k$

\section{Preliminary Experiments}
\label{ap:preliminary_experiments}

\subsection{Preliminary Experiments Setting}
\label{sub:preliminary_exp}

The preliminary experiments are designed to see how the different output layers work without continual learning constraints. This experiment is also conducted to select the learning rate and architecture for further experiments.

We experimented with CIFAR10 and CIFAR100 datasets, and we use resnet20 pre-trained model from here \footnote{\url{https://github.com/chenyaofo/pytorch-cifar-models}} (BSD 3-Clause License). The main idea is to compare those results with our results. We take the frozen pre-trained model and replace the output layer with a new one we train on all data.
(1) We experiment with training on CIFAR10 with a neural network pre-trained on CIFAR100 and the opposite. (2) We also trained on CIFAR10 with the model pre-trained on CIFAR10 to see if we would recover the original accuracy, and we did the same for CIFAR100. Doing continual learning on a dataset with a model pre-trained on this dataset is not recommended since it means having access to the whole data from the beginning of training.  We did not use (2) setting in the further experiments, and it was just used as a reference.

(3) We also experiment also with Core50 dataset \cite{Lomonaco2017core50}. We used models pre-trained on ImageNet available on the torchvision library \cite{NEURIPS2019_9015}. We used VGG16, GoogleNet and Resnet, which have a latent space of dimension 2048, 1024, and 512. We experimented with the Core50 dataset, Core50 setting with 50 classes corresponding to the object id, and Core10 setting with 10 classes corresponding to the object category. 

In these preliminary experiments, we evaluate all layers without their masking variant to verify that the pre-trained models were suited to the continual learning scenarios.

%%%%%%%%%%%%%% CP Past ###############
\subsection{Preliminary Experiments}

The summary of the preliminary results is reported in Table \ref{tab:preliminary_results}. The full results can be found in Table \ref{tab:full_preliminary_results}.  As described in section \ref{sub:preliminary_exp}, we experiment with the simplest output layer for this preliminary experiment, i.e., linear layer, linear layer without bias (Linear\_NB), CosLayer, WeightNorm layer, Original WeightNorm (OWeightNorm), KNN, SLDA, and MeanLayer. 
%%%%%%%%%%%%%%%%%%%%%%%%%%%%%%%%%%%%%%%%%%%%%%%%%%%%%%%%%%%%%%%%%%%

\begin{table*}[!ht]
\centering

  \caption{Preliminary experiments to select learning rate and architectures with existing type of layers. Experiment realized on 8 seeds $[0,1,2,3,4,5,6,7]$.}
  \label{tab:preliminary_results}
  
\resizebox{\textwidth}{!}{
  \begin{tabular}{cccccccccccc}
    \hline 
    Dataset & LR &  Architecture & Pretraining & Linear & Linear\_NB & CosLayer & WeightNorm & OWeightNorm & KNN & SLDA & MeanLayer \\ 
    \hline
    
    CIFAR10 & 
0.1 & 
resnet & 
CIFAR10 & 
$92.55$\mbox{\scriptsize{$\pm0.02$}}& % Linear  
$92.55$\mbox{\scriptsize{$\pm0.02$}}& % Linear_no_bias  
$92.49$\mbox{\scriptsize{$\pm0.05$}}& % CosLayer  
$92.56$\mbox{\scriptsize{$\pm0.03$}}& % WeightNorm  
$92.55$\mbox{\scriptsize{$\pm0.04$}}& % OriginalWeightNorm  
$91.79$\mbox{\scriptsize{$\pm0.26$}}& % KNN  
$92.37$\mbox{\scriptsize{$\pm0.00$}}& % SLDA  
$92.44$\mbox{\scriptsize{$\pm0.00$}}\\% MeanLayer  

\rowcolor{gray!20}CIFAR10 & 
0.1 & 
resnet & 
CIFAR100 & 
$65.37$\mbox{\scriptsize{$\pm2.74$}}& % Linear  
$68.58$\mbox{\scriptsize{$\pm9.55$}}& % Linear_no_bias  
$66.55$\mbox{\scriptsize{$\pm9.83$}}& % CosLayer  
$68.55$\mbox{\scriptsize{$\pm0.21$}}& % WeightNorm  
$69.21$\mbox{\scriptsize{$\pm0.36$}}& % OriginalWeightNorm   
$60.26$\mbox{\scriptsize{$\pm0.28$}}& % KNN  
$66.90$\mbox{\scriptsize{$\pm0.02$}}& % SLDA  
$63.27$\mbox{\scriptsize{$\pm0.00$}}\\% MeanLayer  

CIFAR100 & 
0.1 & 
resnet & 
CIFAR10 & 
$17.32$\mbox{\scriptsize{$\pm0.23$}}& % Linear  
$17.14$\mbox{\scriptsize{$\pm0.22$}}& % Linear_no_bias  
$13.09$\mbox{\scriptsize{$\pm0.22$}}& % CosLayer  
$15.11$\mbox{\scriptsize{$\pm0.22$}}& % WeightNorm  
$16.88$\mbox{\scriptsize{$\pm0.30$}}& % OriginalWeightNorm   
$14.20$\mbox{\scriptsize{$\pm0.17$}}& % KNN  
$26.44$\mbox{\scriptsize{$\pm0.03$}}& % SLDA  
$14.48$\mbox{\scriptsize{$\pm0.00$}}\\% MeanLayer  

\rowcolor{gray!20}CIFAR100 & 
0.1 & 
resnet & 
CIFAR100 & 
$63.97$\mbox{\scriptsize{$\pm0.10$}}& % Linear  
$63.94$\mbox{\scriptsize{$\pm0.11$}}& % Linear_no_bias  
$60.91$\mbox{\scriptsize{$\pm0.24$}}& % CosLayer  
$63.62$\mbox{\scriptsize{$\pm0.09$}}& % WeightNorm  
$63.71$\mbox{\scriptsize{$\pm0.15$}}& % OriginalWeightNorm 
$60.69$\mbox{\scriptsize{$\pm0.12$}}& % KNN  
$61.56$\mbox{\scriptsize{$\pm0.01$}}& % SLDA  
$61.84$\mbox{\scriptsize{$\pm0.00$}}\\% MeanLayer  

 \hline 
Core50 & 
0.1 & 
resnet & 
ImageNet & 
$76.58$\mbox{\scriptsize{$\pm0.41$}}& % Linear  
$76.57$\mbox{\scriptsize{$\pm0.41$}}& % Linear_no_bias  
$53.93$\mbox{\scriptsize{$\pm0.90$}}& % CosLayer  
$77.04$\mbox{\scriptsize{$\pm0.33$}}& % WeightNorm  
$76.85$\mbox{\scriptsize{$\pm0.61$}}& % OriginalWeightNorm  
$65.50$\mbox{\scriptsize{$\pm0.14$}}& % KNN  
$78.55$\mbox{\scriptsize{$\pm0.03$}}& % SLDA  
$71.51$\mbox{\scriptsize{$\pm0.00$}}\\% MeanLayer  
 
\rowcolor{gray!20}Core50 & 
0.1 & 
vgg & 
ImageNet & 
$71.69$\mbox{\scriptsize{$\pm0.50$}}& % Linear  
$71.44$\mbox{\scriptsize{$\pm0.30$}}& % Linear_no_bias  
$61.13$\mbox{\scriptsize{$\pm0.39$}}& % CosLayer  
$72.13$\mbox{\scriptsize{$\pm0.45$}}& % WeightNorm  
$71.53$\mbox{\scriptsize{$\pm0.23$}}& % OriginalWeightNorm  
$56.86$\mbox{\scriptsize{$\pm0.08$}}& % KNN  
$69.98$\mbox{\scriptsize{$\pm0.03$}}& % SLDA  
$66.67$\mbox{\scriptsize{$\pm0.00$}}\\% MeanLayer  

Core50 & 
0.1 & 
googlenet & 
ImageNet & 
$70.13$\mbox{\scriptsize{$\pm0.27$}}& % Linear  
$70.13$\mbox{\scriptsize{$\pm0.27$}}& % Linear_no_bias  
$47.81$\mbox{\scriptsize{$\pm3.26$}}& % CosLayer  
$66.37$\mbox{\scriptsize{$\pm0.74$}}& % WeightNorm  
$69.27$\mbox{\scriptsize{$\pm0.35$}}& % OriginalWeightNorm   
$59.45$\mbox{\scriptsize{$\pm0.11$}}& % KNN  
$70.60$\mbox{\scriptsize{$\pm0.11$}}& % SLDA  
$60.85$\mbox{\scriptsize{$\pm0.15$}}\\% MeanLayer  

\rowcolor{gray!20}Core10Lifelong & 
0.1 & 
resnet & 
ImageNet & 
$85.45$\mbox{\scriptsize{$\pm0.51$}}& % Linear  
$85.62$\mbox{\scriptsize{$\pm0.58$}}& % Linear_no_bias  
$78.76$\mbox{\scriptsize{$\pm0.36$}}& % CosLayer  
$86.29$\mbox{\scriptsize{$\pm0.29$}}& % WeightNorm  
$86.07$\mbox{\scriptsize{$\pm0.51$}}& % OriginalWeightNorm  
$75.71$\mbox{\scriptsize{$\pm1.50$}}& % KNN  
$88.06$\mbox{\scriptsize{$\pm0.01$}}& % SLDA  
$79.61$\mbox{\scriptsize{$\pm0.00$}}\\% MeanLayer  

Core10Lifelong & 
0.1 & 
vgg & 
ImageNet & 
$85.52$\mbox{\scriptsize{$\pm0.57$}}& % Linear  
$85.34$\mbox{\scriptsize{$\pm0.57$}}& % Linear_no_bias  
$79.39$\mbox{\scriptsize{$\pm0.18$}}& % CosLayer  
$85.95$\mbox{\scriptsize{$\pm0.61$}}& % WeightNorm  
$85.56$\mbox{\scriptsize{$\pm0.83$}}& % OriginalWeightNorm  
$75.14$\mbox{\scriptsize{$\pm0.88$}}& % KNN  
$83.69$\mbox{\scriptsize{$\pm0.03$}}& % SLDA  
$78.55$\mbox{\scriptsize{$\pm0.00$}}\\% MeanLayer  

\rowcolor{gray!20}Core10Lifelong & 
0.1 & 
googlenet & 
ImageNet & 
$85.08$\mbox{\scriptsize{$\pm0.47$}}& % Linear  
$85.08$\mbox{\scriptsize{$\pm0.47$}}& % Linear_no_bias  
$78.23$\mbox{\scriptsize{$\pm0.33$}}& % CosLayer  
$83.05$\mbox{\scriptsize{$\pm0.38$}}& % WeightNorm  
$84.97$\mbox{\scriptsize{$\pm0.29$}}& % OriginalWeightNorm  
$77.57$\mbox{\scriptsize{$\pm0.96$}}& % KNN  
$86.33$\mbox{\scriptsize{$\pm0.18$}}& % SLDA  
$76.92$\mbox{\scriptsize{$\pm0.10$}}\\% MeanLayer  

 \hline 

\end{tabular}
}

\end{table*}
%%%%%%%%%%%%%%%%%%%%%%%%%%%%%%%%%%%%%%%%%%%%%%%%%%%%%%%%%%%%%%%%%%%

The results of those preliminary experiments are that globally the learning rate of $0.1$ is adapted to all settings. Secondly, we found that for Core50 and Core10 experiments, the ResNet model was the best suited. Moreover, we see that 5 epochs per task are sufficient to learn a correct solution. 
Concerning CIFAR experiments, the reported top-1 accuracy on CIFAR10 and CIFAR100 of the original pre-trained models are respectively $92.60\%$, $69.83\%$. The settings  CIFAR10 pre-trained on CIFAR10 (same for CIFAR100) show us that the training procedure can recover almost the best performance. In CIFAR100, there is a drop of $6\%$ of accuracy, but it stays reasonable in a 100-way classification task.

Even if not very surprising, one interesting result is that training on CIFAR100 with a model pre-trained on CIFAR10 does not work at all. It illustrates that using a pre-trained model is not always a solution in machine learning or continual learning. They should be selected carefully, even if the data might look similar to CIFAR10 and CIFAR100. On the other hand, the experiment on CIFAR10 with CIFAR100 models shows a significant decrease in the performance for all layers.

These preliminary experiments make us able to fix some hyper-parameters to study all layers in a common setting. It also offers us interesting insight into the drop of performance in simple experiments when using a pre-trained model instead of training end-to-end. The results of this experiment can be used as a baseline for continual experiments and subsets experiments.  

%%%%%%%%%%%%%%%%%%%%%%%%%%%%%%%%%%%%%%%%%%%%%%%%%%%%%%%%%%%%%%%%%%%

\begin{table*}
\centering

  \caption{Preliminary experiments to select learning rate and architectures with existing type of layers}
  \label{tab:full_preliminary_results}
  
\resizebox{\textwidth}{!}{
  \begin{tabular}{cccccccccccc}
    \hline 
    Dataset & LR &  Architecture & Pretraining & Linear & Linear\_NB & CosLayer & WeightNorm & OWeightNorm & KNN & SLDA & MeanLayer \\ 
    \hline
    
    CIFAR10 & 
0.1 & 
resnet & 
CIFAR10 & 
$92.55$\mbox{\scriptsize{$\pm0.02$}}& % Linear  
$92.55$\mbox{\scriptsize{$\pm0.02$}}& % Linear_no_bias  
$92.49$\mbox{\scriptsize{$\pm0.05$}}& % CosLayer  
$92.56$\mbox{\scriptsize{$\pm0.03$}}& % WeightNorm  
$92.55$\mbox{\scriptsize{$\pm0.04$}}& % OriginalWeightNorm  
 - & % KNN  
 - & % SLDA  
 - \\% MeanLayer  
 
\rowcolor{gray!20}CIFAR10 & 
0.01 & 
resnet & 
CIFAR10 & 
$92.59$\mbox{\scriptsize{$\pm0.04$}}& % Linear  
$92.59$\mbox{\scriptsize{$\pm0.03$}}& % Linear_no_bias  
$92.48$\mbox{\scriptsize{$\pm0.03$}}& % CosLayer  
$92.54$\mbox{\scriptsize{$\pm0.02$}}& % WeightNorm  
$92.60$\mbox{\scriptsize{$\pm0.03$}}& % OriginalWeightNorm  
 - & % KNN  
 - & % SLDA  
 - \\% MeanLayer  
 
CIFAR10 & 
0.001 & 
resnet & 
CIFAR10 & 
$92.49$\mbox{\scriptsize{$\pm0.04$}}& % Linear  
$92.49$\mbox{\scriptsize{$\pm0.04$}}& % Linear_no_bias  
$67.03$\mbox{\scriptsize{$\pm10.29$}}& % CosLayer  
$92.39$\mbox{\scriptsize{$\pm0.10$}}& % WeightNorm  
$92.57$\mbox{\scriptsize{$\pm0.04$}}& % OriginalWeightNorm  
 - & % KNN  
 - & % SLDA  
 - \\% MeanLayer  
 
\rowcolor{gray!20}CIFAR10 & 
n/a & 
resnet & 
CIFAR10 & 
 - & % Linear  
 - & % Linear_no_bias  
 - & % CosLayer  
 - & % WeightNorm  
 - & % OriginalWeightNorm  
$91.79$\mbox{\scriptsize{$\pm0.26$}}& % KNN  
$92.37$\mbox{\scriptsize{$\pm0.00$}}& % SLDA  
$92.44$\mbox{\scriptsize{$\pm0.00$}}\\% MeanLayer  

 \hline 
CIFAR10 & 
0.1 & 
resnet & 
CIFAR100 & 
$65.37$\mbox{\scriptsize{$\pm2.74$}}& % Linear  
$68.58$\mbox{\scriptsize{$\pm9.55$}}& % Linear_no_bias  
$66.55$\mbox{\scriptsize{$\pm9.83$}}& % CosLayer  
$68.55$\mbox{\scriptsize{$\pm0.21$}}& % WeightNorm  
$69.21$\mbox{\scriptsize{$\pm0.36$}}& % OriginalWeightNorm  
 - & % KNN  
 - & % SLDA  
 - \\% MeanLayer  
 
\rowcolor{gray!20}CIFAR10 & 
0.01 & 
resnet & 
CIFAR100 & 
$75.13$\mbox{\scriptsize{$\pm10.09$}}& % Linear  
$75.13$\mbox{\scriptsize{$\pm10.08$}}& % Linear_no_bias  
$66.46$\mbox{\scriptsize{$\pm15.03$}}& % CosLayer  
$68.43$\mbox{\scriptsize{$\pm0.08$}}& % WeightNorm  
$69.27$\mbox{\scriptsize{$\pm0.25$}}& % OriginalWeightNorm  
 - & % KNN  
 - & % SLDA  
 - \\% MeanLayer  
 
CIFAR10 & 
0.001 & 
resnet & 
CIFAR100 & 
$73.28$\mbox{\scriptsize{$\pm11.11$}}& % Linear  
$73.30$\mbox{\scriptsize{$\pm11.10$}}& % Linear_no_bias  
$18.85$\mbox{\scriptsize{$\pm12.54$}}& % CosLayer  
$60.71$\mbox{\scriptsize{$\pm0.41$}}& % WeightNorm  
$67.79$\mbox{\scriptsize{$\pm0.14$}}& % OriginalWeightNorm  
 - & % KNN  
 - & % SLDA  
 - \\% MeanLayer  
 
\rowcolor{gray!20}CIFAR10 & 
n/a & 
resnet & 
CIFAR100 & 
 - & % Linear  
 - & % Linear_no_bias  
 - & % CosLayer  
 - & % WeightNorm  
 - & % OriginalWeightNorm  
$60.26$\mbox{\scriptsize{$\pm0.28$}}& % KNN  
$66.90$\mbox{\scriptsize{$\pm0.02$}}& % SLDA  
$63.27$\mbox{\scriptsize{$\pm0.00$}}\\% MeanLayer  

 \hline 
CIFAR100 & 
0.1 & 
resnet & 
CIFAR10 & 
$17.32$\mbox{\scriptsize{$\pm0.23$}}& % Linear  
$17.14$\mbox{\scriptsize{$\pm0.22$}}& % Linear_no_bias  
$13.09$\mbox{\scriptsize{$\pm0.22$}}& % CosLayer  
$15.11$\mbox{\scriptsize{$\pm0.22$}}& % WeightNorm  
$16.88$\mbox{\scriptsize{$\pm0.30$}}& % OriginalWeightNorm  
 - & % KNN  
 - & % SLDA  
 - \\% MeanLayer  
 
\rowcolor{gray!20}CIFAR100 & 
0.01 & 
resnet & 
CIFAR10 & 
$14.86$\mbox{\scriptsize{$\pm0.20$}}& % Linear  
$14.86$\mbox{\scriptsize{$\pm0.18$}}& % Linear_no_bias  
$3.02$\mbox{\scriptsize{$\pm0.43$}}& % CosLayer  
$12.18$\mbox{\scriptsize{$\pm0.28$}}& % WeightNorm  
$15.20$\mbox{\scriptsize{$\pm0.23$}}& % OriginalWeightNorm  
 - & % KNN  
 - & % SLDA  
 - \\% MeanLayer  
 
CIFAR100 & 
0.001 & 
resnet & 
CIFAR10 & 
$6.71$\mbox{\scriptsize{$\pm0.48$}}& % Linear  
$6.71$\mbox{\scriptsize{$\pm0.48$}}& % Linear_no_bias  
$1.22$\mbox{\scriptsize{$\pm0.33$}}& % CosLayer  
$2.40$\mbox{\scriptsize{$\pm0.37$}}& % WeightNorm  
$6.77$\mbox{\scriptsize{$\pm0.44$}}& % OriginalWeightNorm  
 - & % KNN  
 - & % SLDA  
 - \\% MeanLayer  
 
\rowcolor{gray!20}CIFAR100 & 
n/a & 
resnet & 
CIFAR10 & 
 - & % Linear  
 - & % Linear_no_bias  
 - & % CosLayer  
 - & % WeightNorm  
 - & % OriginalWeightNorm  
$14.20$\mbox{\scriptsize{$\pm0.17$}}& % KNN  
$26.44$\mbox{\scriptsize{$\pm0.03$}}& % SLDA  
$14.48$\mbox{\scriptsize{$\pm0.00$}}\\% MeanLayer  

 \hline 
CIFAR100 & 
0.1 & 
resnet & 
CIFAR100 & 
$63.97$\mbox{\scriptsize{$\pm0.10$}}& % Linear  
$63.94$\mbox{\scriptsize{$\pm0.11$}}& % Linear_no_bias  
$60.91$\mbox{\scriptsize{$\pm0.24$}}& % CosLayer  
$63.62$\mbox{\scriptsize{$\pm0.09$}}& % WeightNorm  
$63.71$\mbox{\scriptsize{$\pm0.15$}}& % OriginalWeightNorm  
 - & % KNN  
 - & % SLDA  
 - \\% MeanLayer  
 
\rowcolor{gray!20}CIFAR100 & 
0.01 & 
resnet & 
CIFAR100 & 
$63.70$\mbox{\scriptsize{$\pm0.18$}}& % Linear  
$63.71$\mbox{\scriptsize{$\pm0.19$}}& % Linear_no_bias  
$6.95$\mbox{\scriptsize{$\pm1.19$}}& % CosLayer  
$62.45$\mbox{\scriptsize{$\pm0.26$}}& % WeightNorm  
$63.96$\mbox{\scriptsize{$\pm0.11$}}& % OriginalWeightNorm  
 - & % KNN  
 - & % SLDA  
 - \\% MeanLayer  
 
CIFAR100 & 
0.001 & 
resnet & 
CIFAR100 & 
$54.58$\mbox{\scriptsize{$\pm0.32$}}& % Linear  
$54.54$\mbox{\scriptsize{$\pm0.32$}}& % Linear_no_bias  
$1.29$\mbox{\scriptsize{$\pm0.25$}}& % CosLayer  
$21.35$\mbox{\scriptsize{$\pm1.20$}}& % WeightNorm  
$55.54$\mbox{\scriptsize{$\pm0.48$}}& % OriginalWeightNorm  
 - & % KNN  
 - & % SLDA  
 - \\% MeanLayer  
 
\rowcolor{gray!20}CIFAR100 & 
n/a & 
resnet & 
CIFAR100 & 
 - & % Linear  
 - & % Linear_no_bias  
 - & % CosLayer  
 - & % WeightNorm  
 - & % OriginalWeightNorm  
$60.69$\mbox{\scriptsize{$\pm0.12$}}& % KNN  
$61.56$\mbox{\scriptsize{$\pm0.01$}}& % SLDA  
$61.84$\mbox{\scriptsize{$\pm0.00$}}\\% MeanLayer  

 \hline 

    Core50 & 
0.1 & 
resnet & 
ImageNet & 
$76.58$\mbox{\scriptsize{$\pm0.41$}}& % Linear  
$76.57$\mbox{\scriptsize{$\pm0.41$}}& % Linear_no_bias  
$53.93$\mbox{\scriptsize{$\pm0.90$}}& % CosLayer  
$77.04$\mbox{\scriptsize{$\pm0.33$}}& % WeightNorm  
$76.85$\mbox{\scriptsize{$\pm0.61$}}& % OriginalWeightNorm  
 - & % KNN  
 - & % SLDA  
 - \\% MeanLayer  
 
\rowcolor{gray!20}Core50 & 
0.01 & 
resnet & 
ImageNet & 
$76.29$\mbox{\scriptsize{$\pm0.34$}}& % Linear  
$76.29$\mbox{\scriptsize{$\pm0.34$}}& % Linear_no_bias  
$5.80$\mbox{\scriptsize{$\pm0.89$}}& % CosLayer  
$73.01$\mbox{\scriptsize{$\pm0.51$}}& % WeightNorm  
$76.98$\mbox{\scriptsize{$\pm0.29$}}& % OriginalWeightNorm  
 - & % KNN  
 - & % SLDA  
 - \\% MeanLayer  
 
Core50 & 
0.001 & 
resnet & 
ImageNet & 
$62.64$\mbox{\scriptsize{$\pm0.72$}}& % Linear  
$62.65$\mbox{\scriptsize{$\pm0.72$}}& % Linear_no_bias  
$2.21$\mbox{\scriptsize{$\pm0.56$}}& % CosLayer  
$30.55$\mbox{\scriptsize{$\pm1.05$}}& % WeightNorm  
$63.42$\mbox{\scriptsize{$\pm0.66$}}& % OriginalWeightNorm  
 - & % KNN  
 - & % SLDA  
 - \\% MeanLayer  
 
\rowcolor{gray!20}Core50 & 
n/a & 
resnet & 
ImageNet & 
 - & % Linear  
 - & % Linear_no_bias  
 - & % CosLayer  
 - & % WeightNorm  
 - & % OriginalWeightNorm  
$65.50$\mbox{\scriptsize{$\pm0.14$}}& % KNN  
$78.55$\mbox{\scriptsize{$\pm0.03$}}& % SLDA  
$71.51$\mbox{\scriptsize{$\pm0.00$}}\\% MeanLayer  

 \hline 
Core50 & 
0.1 & 
vgg & 
ImageNet & 
$71.69$\mbox{\scriptsize{$\pm0.50$}}& % Linear  
$71.44$\mbox{\scriptsize{$\pm0.30$}}& % Linear_no_bias  
$61.13$\mbox{\scriptsize{$\pm0.39$}}& % CosLayer  
$72.13$\mbox{\scriptsize{$\pm0.45$}}& % WeightNorm  
$71.53$\mbox{\scriptsize{$\pm0.23$}}& % OriginalWeightNorm  
 - & % KNN  
 - & % SLDA  
 - \\% MeanLayer  
 
\rowcolor{gray!20}Core50 & 
0.01 & 
vgg & 
ImageNet & 
$72.29$\mbox{\scriptsize{$\pm0.28$}}& % Linear  
$72.29$\mbox{\scriptsize{$\pm0.28$}}& % Linear_no_bias  
$27.15$\mbox{\scriptsize{$\pm1.66$}}& % CosLayer  
$71.43$\mbox{\scriptsize{$\pm0.35$}}& % WeightNorm  
$71.91$\mbox{\scriptsize{$\pm0.37$}}& % OriginalWeightNorm  
 - & % KNN  
 - & % SLDA  
 - \\% MeanLayer  
 
Core50 & 
0.001 & 
vgg & 
ImageNet & 
$68.79$\mbox{\scriptsize{$\pm0.32$}}& % Linear  
$68.78$\mbox{\scriptsize{$\pm0.31$}}& % Linear_no_bias  
$3.47$\mbox{\scriptsize{$\pm0.79$}}& % CosLayer  
$59.54$\mbox{\scriptsize{$\pm0.43$}}& % WeightNorm  
$69.06$\mbox{\scriptsize{$\pm0.32$}}& % OriginalWeightNorm  
 - & % KNN  
 - & % SLDA  
 - \\% MeanLayer  
 
\rowcolor{gray!20}Core50 & 
n/a & 
vgg & 
ImageNet & 
 - & % Linear  
 - & % Linear_no_bias  
 - & % CosLayer  
 - & % WeightNorm  
 - & % OriginalWeightNorm  
$56.86$\mbox{\scriptsize{$\pm0.08$}}& % KNN  
$69.98$\mbox{\scriptsize{$\pm0.03$}}& % SLDA  
$66.67$\mbox{\scriptsize{$\pm0.00$}}\\% MeanLayer  

 \hline 
Core50 & 
0.1 & 
googlenet & 
ImageNet & 
$70.13$\mbox{\scriptsize{$\pm0.27$}}& % Linear  
$70.13$\mbox{\scriptsize{$\pm0.27$}}& % Linear_no_bias  
$47.81$\mbox{\scriptsize{$\pm3.26$}}& % CosLayer  
$66.37$\mbox{\scriptsize{$\pm0.74$}}& % WeightNorm  
$69.27$\mbox{\scriptsize{$\pm0.35$}}& % OriginalWeightNorm  
 - & % KNN  
 - & % SLDA  
 - \\% MeanLayer  
 
\rowcolor{gray!20}Core50 & 
0.01 & 
googlenet & 
ImageNet & 
$65.99$\mbox{\scriptsize{$\pm0.50$}}& % Linear  
$65.99$\mbox{\scriptsize{$\pm0.50$}}& % Linear_no_bias  
$8.71$\mbox{\scriptsize{$\pm1.31$}}& % CosLayer  
$61.77$\mbox{\scriptsize{$\pm0.21$}}& % WeightNorm  
$66.65$\mbox{\scriptsize{$\pm0.52$}}& % OriginalWeightNorm  
 - & % KNN  
 - & % SLDA  
 - \\% MeanLayer  
 
Core50 & 
0.001 & 
googlenet & 
ImageNet & 
$49.61$\mbox{\scriptsize{$\pm0.85$}}& % Linear  
$49.62$\mbox{\scriptsize{$\pm0.85$}}& % Linear_no_bias  
$2.55$\mbox{\scriptsize{$\pm0.64$}}& % CosLayer  
$21.91$\mbox{\scriptsize{$\pm1.17$}}& % WeightNorm  
$49.76$\mbox{\scriptsize{$\pm0.66$}}& % OriginalWeightNorm  
 - & % KNN  
 - & % SLDA  
 - \\% MeanLayer  
 
\rowcolor{gray!20}Core50 & 
n/a & 
googlenet & 
ImageNet & 
 - & % Linear  
 - & % Linear_no_bias  
 - & % CosLayer  
 - & % WeightNorm  
 - & % OriginalWeightNorm  
$59.45$\mbox{\scriptsize{$\pm0.11$}}& % KNN  
$70.60$\mbox{\scriptsize{$\pm0.11$}}& % SLDA  
$60.85$\mbox{\scriptsize{$\pm0.15$}}\\% MeanLayer  

 \hline 
Core10Lifelong & 
0.1 & 
resnet & 
ImageNet & 
$85.45$\mbox{\scriptsize{$\pm0.51$}}& % Linear  
$85.62$\mbox{\scriptsize{$\pm0.58$}}& % Linear_no_bias  
$78.76$\mbox{\scriptsize{$\pm0.36$}}& % CosLayer  
$86.29$\mbox{\scriptsize{$\pm0.29$}}& % WeightNorm  
$86.07$\mbox{\scriptsize{$\pm0.51$}}& % OriginalWeightNorm  
 - & % KNN  
 - & % SLDA  
 - \\% MeanLayer  
 
\rowcolor{gray!20}Core10Lifelong & 
0.01 & 
resnet & 
ImageNet & 
$86.19$\mbox{\scriptsize{$\pm0.33$}}& % Linear  
$86.19$\mbox{\scriptsize{$\pm0.33$}}& % Linear_no_bias  
$55.26$\mbox{\scriptsize{$\pm1.84$}}& % CosLayer  
$85.46$\mbox{\scriptsize{$\pm0.40$}}& % WeightNorm  
$86.44$\mbox{\scriptsize{$\pm0.36$}}& % OriginalWeightNorm  
 - & % KNN  
 - & % SLDA  
 - \\% MeanLayer  
 
Core10Lifelong & 
0.001 & 
resnet & 
ImageNet & 
$82.53$\mbox{\scriptsize{$\pm0.52$}}& % Linear  
$82.53$\mbox{\scriptsize{$\pm0.52$}}& % Linear_no_bias  
$13.38$\mbox{\scriptsize{$\pm3.59$}}& % CosLayer  
$76.42$\mbox{\scriptsize{$\pm1.03$}}& % WeightNorm  
$83.41$\mbox{\scriptsize{$\pm0.45$}}& % OriginalWeightNorm  
 - & % KNN  
 - & % SLDA  
 - \\% MeanLayer  
 
\rowcolor{gray!20}Core10Lifelong & 
n/a & 
resnet & 
ImageNet & 
 - & % Linear  
 - & % Linear_no_bias  
 - & % CosLayer  
 - & % WeightNorm  
 - & % OriginalWeightNorm  
$75.71$\mbox{\scriptsize{$\pm1.50$}}& % KNN  
$88.06$\mbox{\scriptsize{$\pm0.01$}}& % SLDA  
$79.61$\mbox{\scriptsize{$\pm0.00$}}\\% MeanLayer  

Core10Lifelong & 
0.1 & 
vgg & 
ImageNet & 
$85.52$\mbox{\scriptsize{$\pm0.57$}}& % Linear  
$85.34$\mbox{\scriptsize{$\pm0.57$}}& % Linear_no_bias  
$79.39$\mbox{\scriptsize{$\pm0.18$}}& % CosLayer  
$85.95$\mbox{\scriptsize{$\pm0.61$}}& % WeightNorm  
$85.56$\mbox{\scriptsize{$\pm0.83$}}& % OriginalWeightNorm  
 - & % KNN  
 - & % SLDA  
 - \\% MeanLayer  
 
\rowcolor{gray!20}Core10Lifelong & 
0.01 & 
vgg & 
ImageNet & 
$85.85$\mbox{\scriptsize{$\pm0.30$}}& % Linear  
$85.89$\mbox{\scriptsize{$\pm0.27$}}& % Linear_no_bias  
$72.44$\mbox{\scriptsize{$\pm1.92$}}& % CosLayer  
$85.37$\mbox{\scriptsize{$\pm0.24$}}& % WeightNorm  
$85.74$\mbox{\scriptsize{$\pm0.65$}}& % OriginalWeightNorm  
 - & % KNN  
 - & % SLDA  
 - \\% MeanLayer  
 
Core10Lifelong & 
0.001 & 
vgg & 
ImageNet & 
$84.20$\mbox{\scriptsize{$\pm0.14$}}& % Linear  
$84.20$\mbox{\scriptsize{$\pm0.14$}}& % Linear_no_bias  
$24.59$\mbox{\scriptsize{$\pm3.17$}}& % CosLayer  
$82.14$\mbox{\scriptsize{$\pm0.47$}}& % WeightNorm  
$84.51$\mbox{\scriptsize{$\pm0.11$}}& % OriginalWeightNorm  
 - & % KNN  
 - & % SLDA  
 - \\% MeanLayer  
 
\rowcolor{gray!20}Core10Lifelong & 
n/a & 
vgg & 
ImageNet & 
 - & % Linear  
 - & % Linear_no_bias  
 - & % CosLayer  
 - & % WeightNorm  
 - & % OriginalWeightNorm  
$75.14$\mbox{\scriptsize{$\pm0.88$}}& % KNN  
$83.69$\mbox{\scriptsize{$\pm0.03$}}& % SLDA  
$78.55$\mbox{\scriptsize{$\pm0.00$}}\\% MeanLayer  

 \hline 
Core10Lifelong & 
0.1 & 
googlenet & 
ImageNet & 
$85.08$\mbox{\scriptsize{$\pm0.47$}}& % Linear  
$85.08$\mbox{\scriptsize{$\pm0.47$}}& % Linear_no_bias  
$78.23$\mbox{\scriptsize{$\pm0.33$}}& % CosLayer  
$83.05$\mbox{\scriptsize{$\pm0.38$}}& % WeightNorm  
$84.97$\mbox{\scriptsize{$\pm0.29$}}& % OriginalWeightNorm  
 - & % KNN  
 - & % SLDA  
 - \\% MeanLayer  
 
\rowcolor{gray!20}Core10Lifelong & 
0.01 & 
googlenet & 
ImageNet & 
$83.67$\mbox{\scriptsize{$\pm0.26$}}& % Linear  
$83.67$\mbox{\scriptsize{$\pm0.25$}}& % Linear_no_bias  
$64.27$\mbox{\scriptsize{$\pm2.53$}}& % CosLayer  
$82.31$\mbox{\scriptsize{$\pm0.28$}}& % WeightNorm  
$83.97$\mbox{\scriptsize{$\pm0.33$}}& % OriginalWeightNorm  
 - & % KNN  
 - & % SLDA  
 - \\% MeanLayer  
 
Core10Lifelong & 
0.001 & 
googlenet & 
ImageNet & 
$79.51$\mbox{\scriptsize{$\pm0.36$}}& % Linear  
$79.52$\mbox{\scriptsize{$\pm0.35$}}& % Linear_no_bias  
$14.85$\mbox{\scriptsize{$\pm2.09$}}& % CosLayer  
$73.90$\mbox{\scriptsize{$\pm0.63$}}& % WeightNorm  
$80.13$\mbox{\scriptsize{$\pm0.30$}}& % OriginalWeightNorm  
 - & % KNN  
 - & % SLDA  
 - \\% MeanLayer  
 
\rowcolor{gray!20}Core10Lifelong & 
n/a & 
googlenet & 
ImageNet & 
 - & % Linear  
 - & % Linear_no_bias  
 - & % CosLayer  
 - & % WeightNorm  
 - & % OriginalWeightNorm  
$77.57$\mbox{\scriptsize{$\pm0.96$}}& % KNN  
$86.33$\mbox{\scriptsize{$\pm0.18$}}& % SLDA  
$76.92$\mbox{\scriptsize{$\pm0.10$}}\\% MeanLayer  

 \hline

\end{tabular}
}

\end{table*}
%%%%%%%%%%%%%%%%%%%%%%%%%%%%%%%%%%%%%%%%%%%%%%%%%%%%%%%%%%%%%%%%%%%

\newpage{}
\section{Forgetting in Lifelong Scenarios}

\subsection{CIFAR100Lifelong}

\begin{figure}[h]
    \centering
    \begin{subfigure}[t]{\linewidth}
        \centering
        \includegraphics[width=\linewidth]{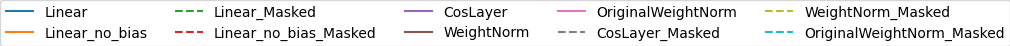}
    \end{subfigure}

    \centering
    \begin{subfigure}[t]{0.75\linewidth}
        \centering
        \includegraphics[width=\linewidth]{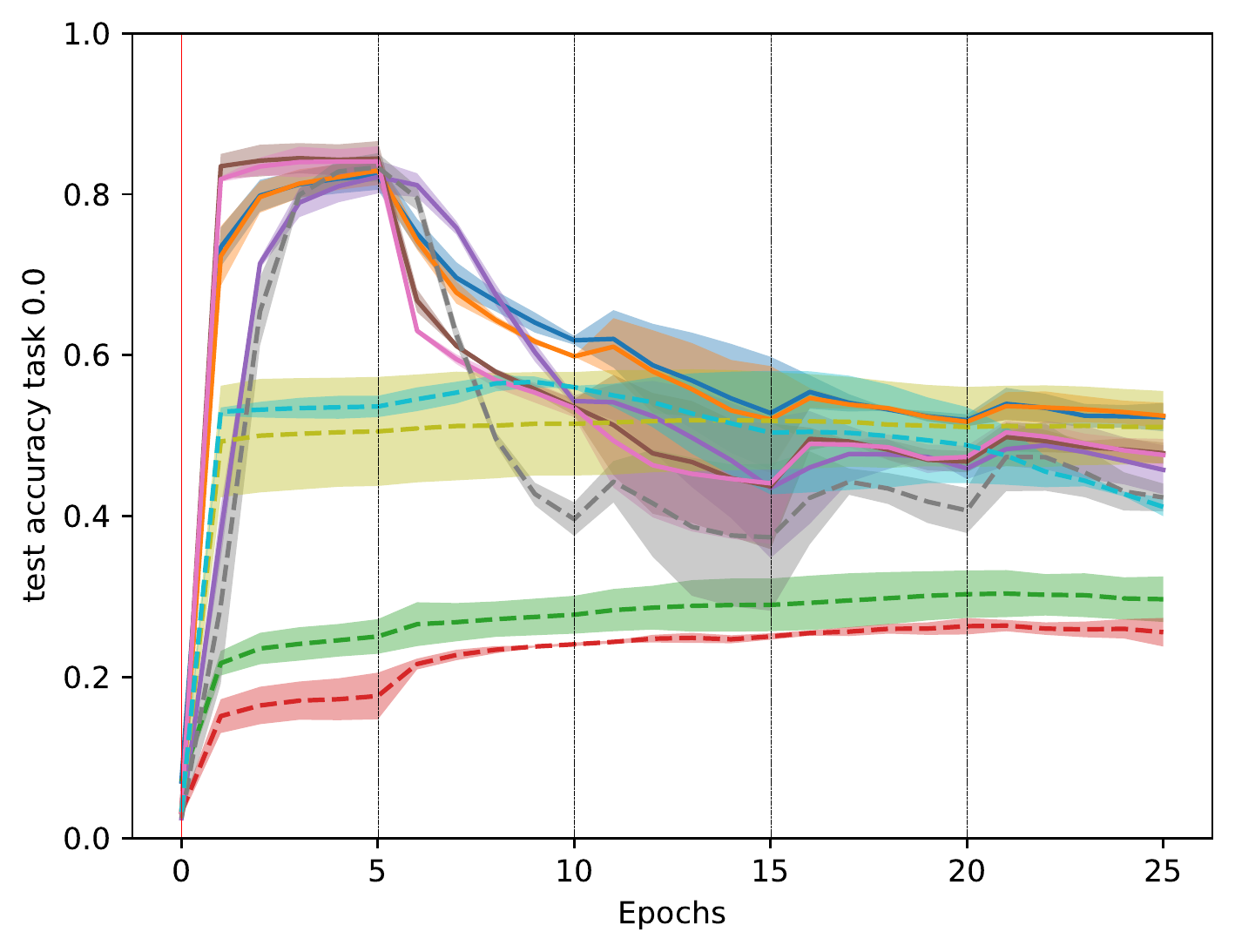}
        \caption{CIFAR10, 5 tasks}
    \label{fig:forget_exp_cifar100life}
    \end{subfigure}
    \caption{Test accuracy on task 0 in CIFAR100Lifelong Scenario.}
    \label{fig:forgetting}
\end{figure}

In Fig.~\ref{fig:forgetting}, we can see that the model is forgetting the first tasks and the masking have no effect on it. Moreover, since the single masking makes learning less efficient with many classes, it can make the performance bad from the beginning.

\subsection{Core10Lifelong}

Unfortunately, the Core50 test set does not make it possible to evaluate the forgetting in each task of lifelong scenarios. Indeed, the environments of the test set are different from the train environments. Then it is not possible to evaluate with the test set the evolution of accuracy on past tasks. In the lifelong scenario, we can only evaluate the performance on 3 other unknown environments, similarly to an out of distribution evaluation. 

\newpage{}
\section{Full subsets results}

%%%%%%%%%%%%%%%%%%%%%%%%%%%%%%%%%%%%%%%%%%%%%%%%%%%%%%%%%%%%%%%%%%%
\begin{table*}
\centering

  \caption{Subset experiments: Mean Accuracy and standard deviation on 8 runs with different seeds.}
  \label{tab:ap:subsets_results}
  
\resizebox{0.75\textwidth}{!}{
  \begin{tabular}{ccccccc}
    \hline 
    OutLayer&Dataset $\backslash$ Subset & 100 & 200 & 500 & 1000 & All\\ 
    \hline

    % ##############  0.1 ############## 
CosLayer & 
CIFAR10 & 
$31.81$\mbox{\scriptsize{$\pm5.84$}}& % 100  
$39.00$\mbox{\scriptsize{$\pm4.90$}}& % 200  
$48.23$\mbox{\scriptsize{$\pm3.51$}}& % 500  
$54.09$\mbox{\scriptsize{$\pm2.66$}}& % 1000  
$66.55$\mbox{\scriptsize{$\pm9.83$}}\\ 
 
\rowcolor{gray!20}WeightNorm & 
CIFAR10 & 
$50.82$\mbox{\scriptsize{$\pm1.91$}}& % 100  
$56.90$\mbox{\scriptsize{$\pm0.82$}}& % 200  
$62.64$\mbox{\scriptsize{$\pm0.78$}}& % 500  
$65.41$\mbox{\scriptsize{$\pm0.49$}}& % 1000  
$68.55$\mbox{\scriptsize{$\pm0.21$}}\\ 
 
OriginalWeightNorm & 
CIFAR10 & 
$48.22$\mbox{\scriptsize{$\pm2.19$}}& % 100  
$52.33$\mbox{\scriptsize{$\pm1.53$}}& % 200  
$55.91$\mbox{\scriptsize{$\pm0.77$}}& % 500  
$61.66$\mbox{\scriptsize{$\pm0.74$}}& % 1000  
$69.21$\mbox{\scriptsize{$\pm0.36$}}\\ 
 
\rowcolor{gray!20}CosLayer-Masked & 
CIFAR10 & 
$53.45$\mbox{\scriptsize{$\pm1.63$}}& % 100  
$58.54$\mbox{\scriptsize{$\pm0.54$}}& % 200  
$61.25$\mbox{\scriptsize{$\pm0.67$}}& % 500  
$62.51$\mbox{\scriptsize{$\pm0.46$}}& % 1000  
$67.36$\mbox{\scriptsize{$\pm9.52$}}\\ 
 
WeightNorm-Masked & 
CIFAR10 & 
$36.55$\mbox{\scriptsize{$\pm3.26$}}& % 100  
$39.04$\mbox{\scriptsize{$\pm2.23$}}& % 200  
$41.23$\mbox{\scriptsize{$\pm2.15$}}& % 500  
$42.44$\mbox{\scriptsize{$\pm1.62$}}& % 1000  
$33.89$\mbox{\scriptsize{$\pm5.33$}}\\ 
 
\rowcolor{gray!20}OriginalWeightNorm-Masked & 
CIFAR10 & 
$22.79$\mbox{\scriptsize{$\pm10.50$}}& % 100  
$22.68$\mbox{\scriptsize{$\pm9.30$}}& % 200  
$21.07$\mbox{\scriptsize{$\pm9.97$}}& % 500  
$22.45$\mbox{\scriptsize{$\pm6.73$}}& % 1000  
$33.18$\mbox{\scriptsize{$\pm3.31$}}\\ 
 
Linear & 
CIFAR10 & 
$49.13$\mbox{\scriptsize{$\pm2.31$}}& % 100  
$54.31$\mbox{\scriptsize{$\pm1.43$}}& % 200  
$60.16$\mbox{\scriptsize{$\pm0.72$}}& % 500  
$64.02$\mbox{\scriptsize{$\pm0.77$}}& % 1000  
$75.13$\mbox{\scriptsize{$\pm10.09$}}\\ 
 
\rowcolor{gray!20}Linear-no-bias & 
CIFAR10 & 
$49.11$\mbox{\scriptsize{$\pm2.27$}}& % 100  
$54.31$\mbox{\scriptsize{$\pm1.42$}}& % 200  
$60.15$\mbox{\scriptsize{$\pm0.73$}}& % 500  
$64.04$\mbox{\scriptsize{$\pm0.77$}}& % 1000  
$75.13$\mbox{\scriptsize{$\pm10.08$}}\\ 
 
Linear-Masked & 
CIFAR10 & 
$23.02$\mbox{\scriptsize{$\pm8.05$}}& % 100  
$25.78$\mbox{\scriptsize{$\pm5.89$}}& % 200  
$27.45$\mbox{\scriptsize{$\pm3.77$}}& % 500  
$30.02$\mbox{\scriptsize{$\pm4.71$}}& % 1000  
$42.57$\mbox{\scriptsize{$\pm29.59$}}\\ 
 
\rowcolor{gray!20}Linear-no-bias-Masked & 
CIFAR10 & 
$23.01$\mbox{\scriptsize{$\pm8.12$}}& % 100  
$25.96$\mbox{\scriptsize{$\pm6.00$}}& % 200  
$27.74$\mbox{\scriptsize{$\pm3.64$}}& % 500  
$30.01$\mbox{\scriptsize{$\pm4.84$}}& % 1000  
$39.20$\mbox{\scriptsize{$\pm30.87$}}\\ 
 
KNN & 
CIFAR10 & 
$45.21$\mbox{\scriptsize{$\pm2.03$}}& % 100  
$49.47$\mbox{\scriptsize{$\pm1.47$}}& % 200  
$54.19$\mbox{\scriptsize{$\pm1.13$}}& % 500  
$57.00$\mbox{\scriptsize{$\pm0.58$}}& % 1000  
$60.26$\mbox{\scriptsize{$\pm0.28$}}\\ 
 
\rowcolor{gray!20}SLDA & 
CIFAR10 & 
$49.23$\mbox{\scriptsize{$\pm22.67$}}& % 100  
$61.51$\mbox{\scriptsize{$\pm16.98$}}& % 200  
$68.47$\mbox{\scriptsize{$\pm13.43$}}& % 500  
$70.47$\mbox{\scriptsize{$\pm12.44$}}& % 1000  
$66.90$\mbox{\scriptsize{$\pm0.02$}}\\ 
 
MeanLayer & 
CIFAR10 & 
$62.34$\mbox{\scriptsize{$\pm17.28$}}& % 100  
$66.58$\mbox{\scriptsize{$\pm14.84$}}& % 200  
$68.65$\mbox{\scriptsize{$\pm13.71$}}& % 500  
$69.58$\mbox{\scriptsize{$\pm13.23$}}& % 1000  
$63.27$\mbox{\scriptsize{$\pm0.00$}}\\ 
 
\rowcolor{gray!20}MedianLayer & 
CIFAR10 & 
$60.65$\mbox{\scriptsize{$\pm18.27$}}& % 100  
$65.50$\mbox{\scriptsize{$\pm15.53$}}& % 200  
$63.59$\mbox{\scriptsize{$\pm10.88$}}& % 500  
$68.84$\mbox{\scriptsize{$\pm13.61$}}& % 1000  
$70.44$\mbox{\scriptsize{$\pm12.72$}}\\

 \hline 
% ##############  0.1 ############## 
CosLayer & 
CIFAR100Lifelong & 
$28.42$\mbox{\scriptsize{$\pm4.38$}}& % 100  
$36.61$\mbox{\scriptsize{$\pm5.10$}}& % 200  
$47.39$\mbox{\scriptsize{$\pm3.01$}}& % 500  
$53.30$\mbox{\scriptsize{$\pm1.87$}}& % 1000  
$65.42$\mbox{\scriptsize{$\pm0.28$}}\\ 
 
\rowcolor{gray!20}WeightNorm & 
CIFAR100Lifelong & 
$49.19$\mbox{\scriptsize{$\pm2.30$}}& % 100  
$57.33$\mbox{\scriptsize{$\pm1.14$}}& % 200  
$63.67$\mbox{\scriptsize{$\pm0.65$}}& % 500  
$66.39$\mbox{\scriptsize{$\pm0.44$}}& % 1000  
$70.52$\mbox{\scriptsize{$\pm0.22$}}\\ 
 
OriginalWeightNorm & 
CIFAR100Lifelong & 
$47.68$\mbox{\scriptsize{$\pm2.52$}}& % 100  
$55.16$\mbox{\scriptsize{$\pm1.10$}}& % 200  
$60.70$\mbox{\scriptsize{$\pm0.61$}}& % 500  
$62.83$\mbox{\scriptsize{$\pm0.69$}}& % 1000  
$70.53$\mbox{\scriptsize{$\pm0.19$}}\\ 
 
\rowcolor{gray!20}CosLayer-Masked & 
CIFAR100Lifelong & 
$53.27$\mbox{\scriptsize{$\pm1.74$}}& % 100  
$58.88$\mbox{\scriptsize{$\pm0.76$}}& % 200  
$62.88$\mbox{\scriptsize{$\pm0.54$}}& % 500  
$64.56$\mbox{\scriptsize{$\pm0.33$}}& % 1000  
$65.74$\mbox{\scriptsize{$\pm0.06$}}\\ 
 
WeightNorm-Masked & 
CIFAR100Lifelong & 
$23.17$\mbox{\scriptsize{$\pm3.88$}}& % 100  
$26.00$\mbox{\scriptsize{$\pm5.85$}}& % 200  
$27.69$\mbox{\scriptsize{$\pm5.08$}}& % 500  
$28.60$\mbox{\scriptsize{$\pm6.01$}}& % 1000  
$30.78$\mbox{\scriptsize{$\pm4.98$}}\\ 
 
\rowcolor{gray!20}OriginalWeightNorm-Masked & 
CIFAR100Lifelong & 
$16.13$\mbox{\scriptsize{$\pm8.01$}}& % 100  
$18.49$\mbox{\scriptsize{$\pm8.51$}}& % 200  
$16.16$\mbox{\scriptsize{$\pm7.56$}}& % 500  
$18.12$\mbox{\scriptsize{$\pm6.86$}}& % 1000  
$24.80$\mbox{\scriptsize{$\pm5.16$}}\\ 
 
Linear & 
CIFAR100Lifelong & 
$47.84$\mbox{\scriptsize{$\pm2.81$}}& % 100  
$55.88$\mbox{\scriptsize{$\pm1.04$}}& % 200  
$62.46$\mbox{\scriptsize{$\pm0.79$}}& % 500  
$65.56$\mbox{\scriptsize{$\pm0.50$}}& % 1000  
$70.36$\mbox{\scriptsize{$\pm0.10$}}\\ 
 
\rowcolor{gray!20}Linear-no-bias & 
CIFAR100Lifelong & 
$47.24$\mbox{\scriptsize{$\pm2.64$}}& % 100  
$55.70$\mbox{\scriptsize{$\pm0.89$}}& % 200  
$62.35$\mbox{\scriptsize{$\pm0.70$}}& % 500  
$65.38$\mbox{\scriptsize{$\pm0.48$}}& % 1000  
$70.28$\mbox{\scriptsize{$\pm0.29$}}\\ 
 
Linear-Masked & 
CIFAR100Lifelong & 
$11.41$\mbox{\scriptsize{$\pm4.22$}}& % 100  
$14.29$\mbox{\scriptsize{$\pm3.81$}}& % 200  
$14.90$\mbox{\scriptsize{$\pm2.85$}}& % 500  
$14.76$\mbox{\scriptsize{$\pm2.52$}}& % 1000  
$16.39$\mbox{\scriptsize{$\pm2.11$}}\\ 
 
\rowcolor{gray!20}Linear-no-bias-Masked & 
CIFAR100Lifelong & 
$11.81$\mbox{\scriptsize{$\pm3.97$}}& % 100  
$12.41$\mbox{\scriptsize{$\pm3.47$}}& % 200  
$12.77$\mbox{\scriptsize{$\pm2.43$}}& % 500  
$13.16$\mbox{\scriptsize{$\pm2.30$}}& % 1000  
$14.53$\mbox{\scriptsize{$\pm2.80$}}\\ 
 
KNN & 
CIFAR100Lifelong & 
$44.78$\mbox{\scriptsize{$\pm2.39$}}& % 100  
$52.98$\mbox{\scriptsize{$\pm1.33$}}& % 200  
$61.87$\mbox{\scriptsize{$\pm0.85$}}& % 500  
$65.93$\mbox{\scriptsize{$\pm0.77$}}& % 1000  
$66.71$\mbox{\scriptsize{$\pm1.01$}}\\ 
 
\rowcolor{gray!20}SLDA & 
CIFAR100Lifelong & 
$37.03$\mbox{\scriptsize{$\pm2.04$}}& % 100  
$55.02$\mbox{\scriptsize{$\pm0.85$}}& % 200  
$63.07$\mbox{\scriptsize{$\pm0.44$}}& % 500  
$65.75$\mbox{\scriptsize{$\pm0.28$}}& % 1000  
$68.17$\mbox{\scriptsize{$\pm0.01$}}\\ 
 
MeanLayer & 
CIFAR100Lifelong & 
$51.63$\mbox{\scriptsize{$\pm2.17$}}& % 100  
$58.01$\mbox{\scriptsize{$\pm1.05$}}& % 200  
$62.32$\mbox{\scriptsize{$\pm0.45$}}& % 500  
$64.13$\mbox{\scriptsize{$\pm0.32$}}& % 1000  
$65.55$\mbox{\scriptsize{$\pm0.00$}}\\ 
 
\rowcolor{gray!20}MedianLayer & 
CIFAR100Lifelong & 
$48.23$\mbox{\scriptsize{$\pm1.74$}}& % 100  
$55.30$\mbox{\scriptsize{$\pm1.46$}}& % 200  
$60.44$\mbox{\scriptsize{$\pm0.35$}}& % 500  
$62.76$\mbox{\scriptsize{$\pm0.37$}}& % 1000  
$64.17$\mbox{\scriptsize{$\pm0.00$}}\\

 \hline 
% ##############  0.1 ############## 
CosLayer & 
Core50 & 
$10.30$\mbox{\scriptsize{$\pm1.98$}}& % 100  
$12.43$\mbox{\scriptsize{$\pm3.56$}}& % 200  
$18.11$\mbox{\scriptsize{$\pm4.17$}}& % 500  
$26.48$\mbox{\scriptsize{$\pm4.16$}}& % 1000  
$53.93$\mbox{\scriptsize{$\pm0.90$}}\\ 
 
\rowcolor{gray!20}WeightNorm & 
Core50 & 
$32.06$\mbox{\scriptsize{$\pm2.93$}}& % 100  
$44.32$\mbox{\scriptsize{$\pm1.92$}}& % 200  
$61.91$\mbox{\scriptsize{$\pm1.10$}}& % 500  
$69.44$\mbox{\scriptsize{$\pm0.67$}}& % 1000  
$77.04$\mbox{\scriptsize{$\pm0.33$}}\\ 
 
OriginalWeightNorm & 
Core50 & 
$33.19$\mbox{\scriptsize{$\pm2.62$}}& % 100  
$45.68$\mbox{\scriptsize{$\pm1.96$}}& % 200  
$62.39$\mbox{\scriptsize{$\pm1.01$}}& % 500  
$69.17$\mbox{\scriptsize{$\pm0.65$}}& % 1000  
$76.85$\mbox{\scriptsize{$\pm0.61$}}\\ 
 
\rowcolor{gray!20}CosLayer-Masked & 
Core50 & 
$30.62$\mbox{\scriptsize{$\pm2.57$}}& % 100  
$41.22$\mbox{\scriptsize{$\pm1.68$}}& % 200  
$56.47$\mbox{\scriptsize{$\pm1.13$}}& % 500  
$63.10$\mbox{\scriptsize{$\pm0.62$}}& % 1000  
$63.43$\mbox{\scriptsize{$\pm5.78$}}\\ 
 
WeightNorm-Masked & 
Core50 & 
$12.95$\mbox{\scriptsize{$\pm2.07$}}& % 100  
$14.98$\mbox{\scriptsize{$\pm2.89$}}& % 200  
$21.13$\mbox{\scriptsize{$\pm1.22$}}& % 500  
$23.42$\mbox{\scriptsize{$\pm2.49$}}& % 1000  
$31.18$\mbox{\scriptsize{$\pm17.50$}}\\ 
 
\rowcolor{gray!20}OriginalWeightNorm-Masked & 
Core50 & 
$12.27$\mbox{\scriptsize{$\pm2.59$}}& % 100  
$17.42$\mbox{\scriptsize{$\pm2.52$}}& % 200  
$20.82$\mbox{\scriptsize{$\pm4.94$}}& % 500  
$22.21$\mbox{\scriptsize{$\pm5.50$}}& % 1000  
$22.43$\mbox{\scriptsize{$\pm4.36$}}\\ 
 
Linear & 
Core50 & 
$32.80$\mbox{\scriptsize{$\pm2.67$}}& % 100  
$44.88$\mbox{\scriptsize{$\pm1.94$}}& % 200  
$61.80$\mbox{\scriptsize{$\pm1.06$}}& % 500  
$69.03$\mbox{\scriptsize{$\pm0.65$}}& % 1000  
$76.29$\mbox{\scriptsize{$\pm0.34$}}\\ 
 
\rowcolor{gray!20}Linear-no-bias & 
Core50 & 
$32.80$\mbox{\scriptsize{$\pm2.68$}}& % 100  
$44.89$\mbox{\scriptsize{$\pm1.94$}}& % 200  
$61.80$\mbox{\scriptsize{$\pm1.06$}}& % 500  
$69.02$\mbox{\scriptsize{$\pm0.65$}}& % 1000  
$76.29$\mbox{\scriptsize{$\pm0.34$}}\\ 
 
Linear-Masked & 
Core50 & 
$7.37$\mbox{\scriptsize{$\pm1.98$}}& % 100  
$8.20$\mbox{\scriptsize{$\pm2.68$}}& % 200  
$14.02$\mbox{\scriptsize{$\pm5.88$}}& % 500  
$18.46$\mbox{\scriptsize{$\pm2.60$}}& % 1000  
$27.25$\mbox{\scriptsize{$\pm28.39$}}\\ 
 
\rowcolor{gray!20}Linear-no-bias-Masked & 
Core50 & 
$7.36$\mbox{\scriptsize{$\pm1.97$}}& % 100  
$8.22$\mbox{\scriptsize{$\pm2.67$}}& % 200  
$14.04$\mbox{\scriptsize{$\pm5.88$}}& % 500  
$18.48$\mbox{\scriptsize{$\pm2.55$}}& % 1000  
$25.81$\mbox{\scriptsize{$\pm29.13$}}\\ 
 
KNN & 
Core50 & 
$25.95$\mbox{\scriptsize{$\pm3.09$}}& % 100  
$34.17$\mbox{\scriptsize{$\pm2.14$}}& % 200  
$45.50$\mbox{\scriptsize{$\pm2.04$}}& % 500  
$52.13$\mbox{\scriptsize{$\pm1.19$}}& % 1000  
$65.50$\mbox{\scriptsize{$\pm0.14$}}\\ 
 
\rowcolor{gray!20}SLDA & 
Core50 & 
$17.30$\mbox{\scriptsize{$\pm1.23$}}& % 100  
$30.23$\mbox{\scriptsize{$\pm5.58$}}& % 200  
$17.71$\mbox{\scriptsize{$\pm1.47$}}& % 500  
$60.14$\mbox{\scriptsize{$\pm0.93$}}& % 1000  
$78.55$\mbox{\scriptsize{$\pm0.03$}}\\ 
 
MeanLayer & 
Core50 & 
$28.81$\mbox{\scriptsize{$\pm2.68$}}& % 100  
$40.71$\mbox{\scriptsize{$\pm1.65$}}& % 200  
$56.52$\mbox{\scriptsize{$\pm1.29$}}& % 500  
$63.20$\mbox{\scriptsize{$\pm0.79$}}& % 1000  
$71.51$\mbox{\scriptsize{$\pm0.00$}}\\ 
 
\rowcolor{gray!20}MedianLayer & 
Core50 & 
$26.83$\mbox{\scriptsize{$\pm2.26$}}& % 100  
$36.03$\mbox{\scriptsize{$\pm1.65$}}& % 200  
$53.07$\mbox{\scriptsize{$\pm1.15$}}& % 500  
$60.73$\mbox{\scriptsize{$\pm0.77$}}& % 1000  
$70.22$\mbox{\scriptsize{$\pm0.00$}}\\

 \hline 
% ##############  0.1 ############## 
CosLayer & 
Core10Lifelong & 
$31.45$\mbox{\scriptsize{$\pm10.30$}}& % 100  
$43.25$\mbox{\scriptsize{$\pm9.15$}}& % 200  
$52.17$\mbox{\scriptsize{$\pm10.29$}}& % 500  
$65.40$\mbox{\scriptsize{$\pm5.68$}}& % 1000  
$78.76$\mbox{\scriptsize{$\pm0.36$}}\\ 
 
\rowcolor{gray!20}WeightNorm & 
Core10Lifelong & 
$68.52$\mbox{\scriptsize{$\pm4.46$}}& % 100  
$76.78$\mbox{\scriptsize{$\pm1.44$}}& % 200  
$81.38$\mbox{\scriptsize{$\pm0.89$}}& % 500  
$83.89$\mbox{\scriptsize{$\pm0.62$}}& % 1000  
$86.29$\mbox{\scriptsize{$\pm0.29$}}\\ 
 
OriginalWeightNorm & 
Core10Lifelong & 
$67.50$\mbox{\scriptsize{$\pm5.30$}}& % 100  
$75.86$\mbox{\scriptsize{$\pm1.15$}}& % 200  
$80.33$\mbox{\scriptsize{$\pm0.89$}}& % 500  
$82.42$\mbox{\scriptsize{$\pm0.75$}}& % 1000  
$86.07$\mbox{\scriptsize{$\pm0.51$}}\\ 
 
\rowcolor{gray!20}CosLayer-Masked & 
Core10Lifelong & 
$65.58$\mbox{\scriptsize{$\pm4.18$}}& % 100  
$72.28$\mbox{\scriptsize{$\pm2.19$}}& % 200  
$76.89$\mbox{\scriptsize{$\pm1.28$}}& % 500  
$78.69$\mbox{\scriptsize{$\pm0.66$}}& % 1000  
$79.56$\mbox{\scriptsize{$\pm0.65$}}\\ 
 
WeightNorm-Masked & 
Core10Lifelong & 
$42.05$\mbox{\scriptsize{$\pm5.58$}}& % 100  
$48.26$\mbox{\scriptsize{$\pm4.63$}}& % 200  
$54.12$\mbox{\scriptsize{$\pm6.72$}}& % 500  
$57.78$\mbox{\scriptsize{$\pm4.81$}}& % 1000  
$58.74$\mbox{\scriptsize{$\pm16.05$}}\\ 
 
\rowcolor{gray!20}OriginalWeightNorm-Masked & 
Core10Lifelong & 
$26.76$\mbox{\scriptsize{$\pm5.72$}}& % 100  
$34.75$\mbox{\scriptsize{$\pm5.45$}}& % 200  
$30.12$\mbox{\scriptsize{$\pm11.94$}}& % 500  
$31.73$\mbox{\scriptsize{$\pm7.00$}}& % 1000  
$46.05$\mbox{\scriptsize{$\pm4.96$}}\\ 
 
Linear & 
Core10Lifelong & 
$68.47$\mbox{\scriptsize{$\pm4.57$}}& % 100  
$76.33$\mbox{\scriptsize{$\pm1.32$}}& % 200  
$80.79$\mbox{\scriptsize{$\pm0.91$}}& % 500  
$83.08$\mbox{\scriptsize{$\pm0.57$}}& % 1000  
$86.19$\mbox{\scriptsize{$\pm0.33$}}\\ 
 
\rowcolor{gray!20}Linear-no-bias & 
Core10Lifelong & 
$68.47$\mbox{\scriptsize{$\pm4.57$}}& % 100  
$76.33$\mbox{\scriptsize{$\pm1.32$}}& % 200  
$80.79$\mbox{\scriptsize{$\pm0.91$}}& % 500  
$83.08$\mbox{\scriptsize{$\pm0.57$}}& % 1000  
$86.19$\mbox{\scriptsize{$\pm0.33$}}\\ 
 
Linear-Masked & 
Core10Lifelong & 
$23.17$\mbox{\scriptsize{$\pm6.06$}}& % 100  
$24.29$\mbox{\scriptsize{$\pm6.54$}}& % 200  
$29.25$\mbox{\scriptsize{$\pm7.45$}}& % 500  
$30.79$\mbox{\scriptsize{$\pm10.56$}}& % 1000  
$38.08$\mbox{\scriptsize{$\pm28.04$}}\\ 
 
\rowcolor{gray!20}Linear-no-bias-Masked & 
Core10Lifelong & 
$23.11$\mbox{\scriptsize{$\pm5.98$}}& % 100  
$24.06$\mbox{\scriptsize{$\pm6.49$}}& % 200  
$28.82$\mbox{\scriptsize{$\pm7.34$}}& % 500  
$30.76$\mbox{\scriptsize{$\pm10.48$}}& % 1000  
$38.27$\mbox{\scriptsize{$\pm28.03$}}\\ 
 
KNN & 
Core10Lifelong & 
$55.05$\mbox{\scriptsize{$\pm4.00$}}& % 100  
$64.80$\mbox{\scriptsize{$\pm2.23$}}& % 200  
$72.20$\mbox{\scriptsize{$\pm2.05$}}& % 500  
$76.51$\mbox{\scriptsize{$\pm1.55$}}& % 1000  
$75.71$\mbox{\scriptsize{$\pm1.50$}}\\ 
 
\rowcolor{gray!20}SLDA & 
Core10Lifelong & 
$63.00$\mbox{\scriptsize{$\pm3.03$}}& % 100  
$63.96$\mbox{\scriptsize{$\pm2.35$}}& % 200  
$35.55$\mbox{\scriptsize{$\pm1.38$}}& % 500  
$73.46$\mbox{\scriptsize{$\pm0.82$}}& % 1000  
$88.06$\mbox{\scriptsize{$\pm0.01$}}\\ 
 
MeanLayer & 
Core10Lifelong & 
$65.06$\mbox{\scriptsize{$\pm4.08$}}& % 100  
$71.30$\mbox{\scriptsize{$\pm2.70$}}& % 200  
$76.25$\mbox{\scriptsize{$\pm1.37$}}& % 500  
$78.33$\mbox{\scriptsize{$\pm0.49$}}& % 1000  
$79.61$\mbox{\scriptsize{$\pm0.00$}}\\ 
 
\rowcolor{gray!20}MedianLayer & 
Core10Lifelong & 
$61.46$\mbox{\scriptsize{$\pm4.26$}}& % 100  
$69.28$\mbox{\scriptsize{$\pm2.51$}}& % 200  
$74.81$\mbox{\scriptsize{$\pm1.24$}}& % 500  
$76.69$\mbox{\scriptsize{$\pm0.85$}}& % 1000  
$78.16$\mbox{\scriptsize{$\pm0.03$}}\\

 \hline 
% ##############  0.1 ############## 
CosLayer & 
CUB200 & 
$5.19$\mbox{\scriptsize{$\pm0.77$}}& % 100  
$5.41$\mbox{\scriptsize{$\pm1.01$}}& % 200  
$6.77$\mbox{\scriptsize{$\pm1.19$}}& % 500  
$8.74$\mbox{\scriptsize{$\pm1.94$}}& % 1000  
$30.00$\mbox{\scriptsize{$\pm0.73$}}\\ 
 
\rowcolor{gray!20}WeightNorm & 
CUB200 & 
$11.13$\mbox{\scriptsize{$\pm1.19$}}& % 100  
$15.78$\mbox{\scriptsize{$\pm0.83$}}& % 200  
$25.39$\mbox{\scriptsize{$\pm0.98$}}& % 500  
$34.96$\mbox{\scriptsize{$\pm0.93$}}& % 1000  
$54.41$\mbox{\scriptsize{$\pm0.50$}}\\ 
 
OriginalWeightNorm & 
CUB200 & 
$11.17$\mbox{\scriptsize{$\pm0.87$}}& % 100  
$16.16$\mbox{\scriptsize{$\pm0.96$}}& % 200  
$26.10$\mbox{\scriptsize{$\pm0.93$}}& % 500  
$35.50$\mbox{\scriptsize{$\pm0.57$}}& % 1000  
$52.90$\mbox{\scriptsize{$\pm0.67$}}\\ 
 
\rowcolor{gray!20}CosLayer-Masked & 
CUB200 & 
$11.43$\mbox{\scriptsize{$\pm1.20$}}& % 100  
$16.85$\mbox{\scriptsize{$\pm0.41$}}& % 200  
$26.79$\mbox{\scriptsize{$\pm0.77$}}& % 500  
$36.03$\mbox{\scriptsize{$\pm0.74$}}& % 1000  
$33.07$\mbox{\scriptsize{$\pm0.63$}}\\ 
 
WeightNorm-Masked & 
CUB200 & 
$7.51$\mbox{\scriptsize{$\pm1.02$}}& % 100  
$9.10$\mbox{\scriptsize{$\pm1.04$}}& % 200  
$12.37$\mbox{\scriptsize{$\pm1.37$}}& % 500  
$14.40$\mbox{\scriptsize{$\pm0.92$}}& % 1000  
$16.24$\mbox{\scriptsize{$\pm0.85$}}\\ 
 
\rowcolor{gray!20}OriginalWeightNorm-Masked & 
CUB200 & 
$3.96$\mbox{\scriptsize{$\pm1.02$}}& % 100  
$3.92$\mbox{\scriptsize{$\pm0.82$}}& % 200  
$4.39$\mbox{\scriptsize{$\pm0.74$}}& % 500  
$4.96$\mbox{\scriptsize{$\pm1.26$}}& % 1000  
$5.75$\mbox{\scriptsize{$\pm1.02$}}\\ 
 
Linear & 
CUB200 & 
$10.88$\mbox{\scriptsize{$\pm0.82$}}& % 100  
$15.38$\mbox{\scriptsize{$\pm0.86$}}& % 200  
$24.27$\mbox{\scriptsize{$\pm0.86$}}& % 500  
$33.29$\mbox{\scriptsize{$\pm0.93$}}& % 1000  
$28.84$\mbox{\scriptsize{$\pm0.75$}}\\ 
 
\rowcolor{gray!20}Linear-no-bias & 
CUB200 & 
$10.73$\mbox{\scriptsize{$\pm0.82$}}& % 100  
$15.07$\mbox{\scriptsize{$\pm0.73$}}& % 200  
$24.23$\mbox{\scriptsize{$\pm0.86$}}& % 500  
$33.49$\mbox{\scriptsize{$\pm0.91$}}& % 1000  
$28.92$\mbox{\scriptsize{$\pm0.78$}}\\ 
 
Linear-Masked & 
CUB200 & 
$2.07$\mbox{\scriptsize{$\pm0.53$}}& % 100  
$2.53$\mbox{\scriptsize{$\pm0.57$}}& % 200  
$2.55$\mbox{\scriptsize{$\pm0.55$}}& % 500  
$2.77$\mbox{\scriptsize{$\pm0.42$}}& % 1000  
$2.24$\mbox{\scriptsize{$\pm0.24$}}\\ 
 
\rowcolor{gray!20}Linear-no-bias-Masked & 
CUB200 & 
$2.42$\mbox{\scriptsize{$\pm0.33$}}& % 100  
$2.15$\mbox{\scriptsize{$\pm0.43$}}& % 200  
$2.36$\mbox{\scriptsize{$\pm0.59$}}& % 500  
$2.72$\mbox{\scriptsize{$\pm0.58$}}& % 1000  
$2.08$\mbox{\scriptsize{$\pm0.29$}}\\ 
 
KNN & 
CUB200 & 
$10.33$\mbox{\scriptsize{$\pm1.22$}}& % 100  
$14.31$\mbox{\scriptsize{$\pm0.56$}}& % 200  
$20.41$\mbox{\scriptsize{$\pm1.07$}}& % 500  
$25.54$\mbox{\scriptsize{$\pm0.75$}}& % 1000  
$43.79$\mbox{\scriptsize{$\pm0.00$}}\\ 
 
\rowcolor{gray!20}SLDA & 
CUB200 & 
$1.94$\mbox{\scriptsize{$\pm0.65$}}& % 100  
$2.37$\mbox{\scriptsize{$\pm0.70$}}& % 200  
$3.73$\mbox{\scriptsize{$\pm0.44$}}& % 500  
$15.93$\mbox{\scriptsize{$\pm2.68$}}& % 1000  
$58.61$\mbox{\scriptsize{$\pm0.12$}}\\ 
 
MeanLayer & 
CUB200 & 
$10.37$\mbox{\scriptsize{$\pm1.25$}}& % 100  
$15.07$\mbox{\scriptsize{$\pm0.46$}}& % 200  
$25.24$\mbox{\scriptsize{$\pm1.05$}}& % 500  
$34.55$\mbox{\scriptsize{$\pm0.74$}}& % 1000  
$54.14$\mbox{\scriptsize{$\pm0.00$}}\\ 
 
\rowcolor{gray!20}MedianLayer & 
CUB200 & 
$10.77$\mbox{\scriptsize{$\pm1.02$}}& % 100  
$14.92$\mbox{\scriptsize{$\pm0.54$}}& % 200  
$23.07$\mbox{\scriptsize{$\pm0.90$}}& % 500  
$31.91$\mbox{\scriptsize{$\pm1.00$}}& % 1000  
$53.28$\mbox{\scriptsize{$\pm0.00$}}\\

 \hline 

\end{tabular}
}

\end{table*}
%%%%%%%%%%%%%%%%%%%%%%%%%%%%%%%%%%%%%%%%%%%%%%%%%%%%%%%%%%%%%%%%%%%

\newpage{}

\section{Visualizations}
\label{ap:visualization}

\begin{figure}[!ht]

    \centering
    \begin{subfigure}[t]{0.3\linewidth}
        \centering
        \includegraphics[width=\linewidth]{main/Images/Visualization/Linear_angles_between_vectors.png}
        \label{fig:ap:linear_vectors_angles}
        \caption{Linear: Angles between the different vectors of the output layer}
    \end{subfigure}
    \begin{subfigure}[t]{0.3\linewidth}
        \centering
        \includegraphics[width=\linewidth]{main/Images/Visualization/Linear_angles_between_vectors_and_data.png}
        \label{fig:ap:linear_vectors_angles_data}
        \caption{Linear: Mean Angles between the vectors  and data class by class.}
    \end{subfigure}
    \begin{subfigure}[t]{0.3\linewidth}
        \centering
        \includegraphics[width=\linewidth]{main/Images/Visualization/Linear_interference_risks.png}
        \label{fig:ap:linear_interference_risks}
        \caption{Linear: Interference Risks}
    \end{subfigure}

    \begin{subfigure}[t]{0.3\linewidth}
        \centering
        \includegraphics[width=\linewidth]{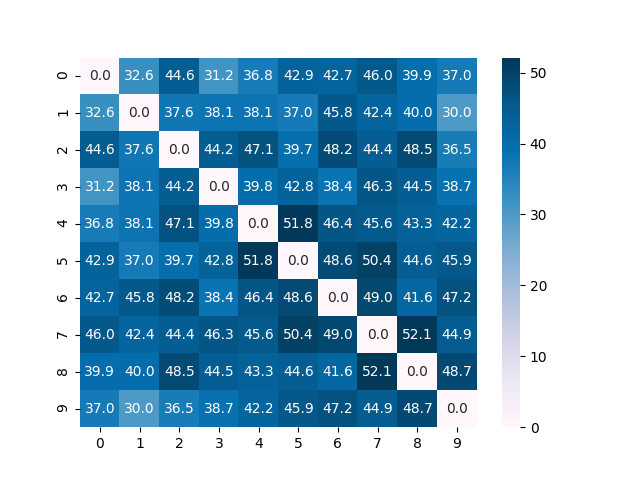}
        \label{fig:ap:weightnorm_vectors_angles}
        \caption{WeightNorm: Angles between the different vectors of the output layer}
    \end{subfigure}
    \begin{subfigure}[t]{0.3\linewidth}
        \centering
        \includegraphics[width=\linewidth]{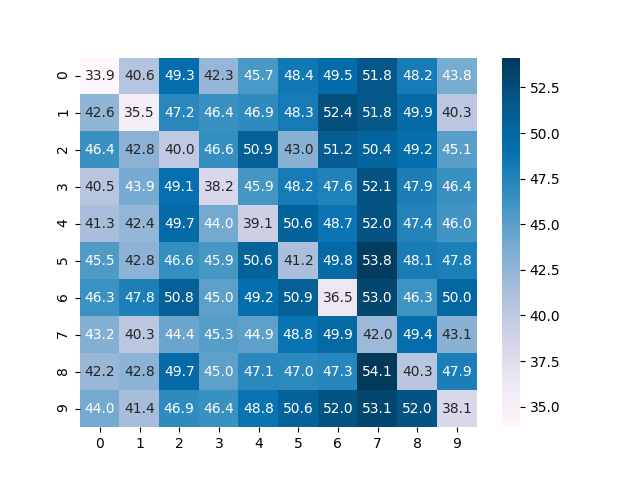}
        \label{fig:ap:weightnorm_vectors_angles_data}
        \caption{WeightNorm: Mean Angles between the vectors  and data class by class.}
    \end{subfigure}
    \begin{subfigure}[t]{0.3\linewidth}
        \centering
        \includegraphics[width=\linewidth]{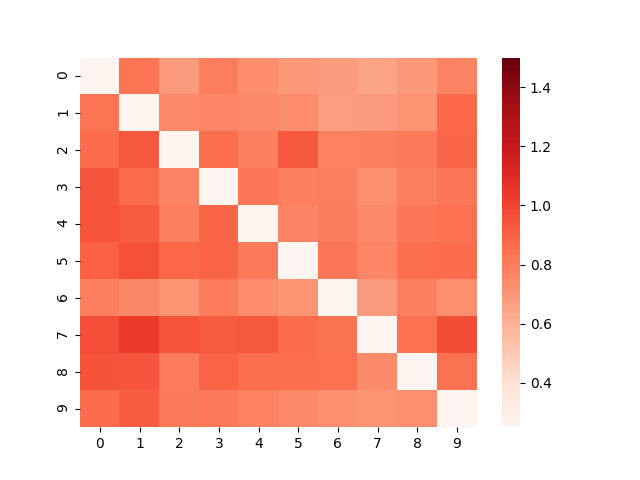}
        \label{fig:ap:weightnorm_interference_risks}
        \caption{WeightNorm: Interference Risks}
    \end{subfigure}
    
\label{fig:ap:visualization}
\end{figure}

\section{Task orders}
\label{sub:task_orders}

\begin{figure}[!ht]
     \begin{subfigure}[b]{\textwidth}
         \begin{python}
         from continuum.scenarios import create_subscenario
         import numpy as np
         
        # we suppose scenario, num_tasks and seed already set
        np.random.seed(seed)
        task_order = np.arange(num_tasks)
        scenario = create_subscenario(scenario, task_order)
        
        \end{python}
         \caption{Code to change task order of the original scenario. We used seeds $[0,1,2,3,4,5,6,7]$. Note: for seed 0, we use original order without modification.}
     \end{subfigure}
\end{figure}

\newpage
\section{Scenario reproducibilty}
\label{sub:scenario_repro}

\begin{figure}[!ht]
     \hfill
     \begin{subfigure}[b]{0.9\textwidth}
         \begin{python}
        from continuum.datasets import Core50
        from continuum import ClassIncremental
        
        dataset = dataset = Core50("./datasets",
                                    download=True,
                                    train=train)
        scenario = ClassIncremental(dataset,
                                    nb_tasks=10,
                                    transformations=transform)
        
        \end{python}
         \caption{Code for Core50 scenario}
     \end{subfigure}

     \hfill
     \begin{subfigure}[b]{0.9\textwidth}
        %\begin{lstlisting}
        \begin{python}
        from continuum.datasets import Core50
        from continuum import ContinualScenario
        
        dataset = Core50("./datasets",
                            scenario="domains",
                            classification="category",
                            train=train)
        scenario = ContinualScenario(dataset,
                                    transformations=transform)
        
        \end{python}
         \caption{Code for Core10Lifelong scenario}
     \end{subfigure}
     
     \hfill
     \begin{subfigure}[b]{0.9\textwidth}
         \begin{python}
        from continuum.datasets import Core50
        from continuum import ContinualScenario
        
        dataset = Core50("./datasets",
                        scenario="objects",
                        classification="category",
                        train=train)
        scenario = ContinualScenario(dataset,
                                    transformations=transform)
        
        \end{python}
         \caption{Code for Core10Mix scenario}
     \end{subfigure}

     \hfill
     \begin{subfigure}[b]{0.9\textwidth}
         \begin{python}
        from continuum.datasets import CUB200
        from continuum import ClassIncremental
        
        dataset = CUB200("./datasets", 
                        train=train)
        scenario = ClassIncremental(dataset,
                                    nb_tasks=10,
                                    transformations=transform)
        
        \end{python}
         \caption{Code for CUB200 scenario}
     \end{subfigure}

     \hfill
     \begin{subfigure}[b]{0.9\textwidth}
         \begin{python}
        from continuum.datasets import CIFAR100
        from continuum import ContinualScenario
        
        dataset = CIFAR100("./datasets",
                        labels_type="category",
                        task_labels="lifelong",
                        train=train)
        scenario = ContinualScenario(dataset, 
                                    transformations=transform)
        
        \end{python}
         \caption{Code for CIFAR100Lifelong scenario}
     \end{subfigure}
\caption{Code to reproduce the scenarios used in the paper with continuum library.}
\label{fig:code_scenarios}
\end{figure}

\end{document}